\RequirePackage{fix-cm}
\documentclass[10pt,twocolumn,letterpaper]{article}
\usepackage[pagenumbers]{cvpr}
\makeatletter
\edef\smRestoreLayout{%
  \textfloatsep=\the\textfloatsep\relax \dbltextfloatsep=\the\dbltextfloatsep\relax
  \floatsep=\the\floatsep\relax \dblfloatsep=\the\dblfloatsep\relax \intextsep=\the\intextsep\relax
  \clubpenalty=\the\clubpenalty\relax \widowpenalty=\the\widowpenalty\relax \brokenpenalty=\the\brokenpenalty\relax
  \c@topnumber=\the\c@topnumber\relax \c@dbltopnumber=\the\c@dbltopnumber\relax \c@totalnumber=\the\c@totalnumber\relax
  \def\noexpand\topfraction{\topfraction}\def\noexpand\dbltopfraction{\dbltopfraction}\def\noexpand\textfraction{\textfraction}%
  \def\noexpand\floatpagefraction{\floatpagefraction}\def\noexpand\dblfloatpagefraction{.85}}
\makeatother
\usepackage{graphicx}
\usepackage{amsmath}
\usepackage{amssymb}
\usepackage{booktabs}
\usepackage{stfloats}
\usepackage{placeins}
\usepackage{needspace}
\usepackage{float}

\usepackage{array}
\definecolor{NBbestbg}{HTML}{F2C5C0}
\definecolor{NBsecondbg}{HTML}{F7D9B5}
\newcommand{\NBbest}[1]{{\setlength{\fboxsep}{0.7pt}\colorbox{NBbestbg}{\strut\textbf{#1}}}}
\newcommand{\NBsecond}[1]{{\setlength{\fboxsep}{0.7pt}\colorbox{NBsecondbg}{\strut#1}}}
\newcommand{\NBplain}[1]{\hspace{0.7pt}\strut#1\hspace{0.7pt}}
\newcommand{\NBbold}[1]{\NBplain{\textbf{#1}}}
\newcommand{\NBmissing}{\NBplain{\textemdash}}
\newcommand{\NBmetric}[1]{\textbf{#1}\,$\downarrow$}
\newcommand{\NBenabled}{\ensuremath{\checkmark}}
\newcommand{\NBdataset}[2]{\multicolumn{5}{@{}l@{}}{\textbf{\textit{#1}}\enspace #2}\\}
\renewcommand{\topfraction}{0.95}
\renewcommand{\dbltopfraction}{0.95}
\renewcommand{\textfraction}{0.05}
\renewcommand{\floatpagefraction}{0.85}
\renewcommand{\dblfloatpagefraction}{0.85}
\usepackage{microtype}
\renewcommand{\topfraction}{1.0}
\renewcommand{\dbltopfraction}{1.0}
\renewcommand{\textfraction}{0.0}

\usepackage{tabularx,multirow,algpseudocode,ragged2e}
\DeclareSymbolFont{smDigits}{OT1}{ptm}{m}{n}
\makeatletter
\newcommand{\smTableNumbers}{%
  \@tempcnta=48\relax
  \loop\mathcode\@tempcnta=\numexpr256*\symsmDigits+\@tempcnta\relax
    \advance\@tempcnta by 1\relax\ifnum\@tempcnta<58 \repeat
  \mathcode46=\numexpr256*\symsmDigits+46\relax
  \mathcode44=\numexpr"6000+256*\symsmDigits+44\relax}
\makeatother
\newcommand{\smTableBody}{%
  \normalfont\fontsize{8}{9.6}\selectfont
  \setlength{\tabcolsep}{1.8pt}%
  \setlength{\aboverulesep}{1pt}\setlength{\belowrulesep}{1pt}%
  \renewcommand{\arraystretch}{1.12}}
\newcommand{\smWideTableBody}{%
  \smTableBody\fontsize{8.2}{9.6}\selectfont
  \setlength{\tabcolsep}{3pt}\renewcommand{\arraystretch}{1.08}}

\newsavebox{\panelart}
\newsavebox{\smNewRealmArt}
\newcommand{\smRealmReplacement}[3][]{\begingroup
  \sbox{\panelart}{\rule{\linewidth}{#2\linewidth}}%
  \sbox{\smNewRealmArt}{\includegraphics[#1,width=\wd\panelart,height=\ht\panelart,keepaspectratio]{#3}}%
  \makebox[\wd\panelart][c]{\raisebox{\dimexpr(\ht\panelart-\ht\smNewRealmArt)/2\relax}%
    [\ht\panelart][\dp\panelart]{\usebox{\smNewRealmArt}}}\endgroup}
\pdfmapline{+smnoto NotoSerifSC-SemiBold "SMNotoEncoding ReEncodeFont" <smnoto.enc <smnoto.ttf}
\font\smFigureFace=smnoto at 7.1601bp
\let\smFigureSubFace\smFigureFace
\newsavebox{\smOldLabel}
\newcommand{\smSameLabel}[2]{\begingroup
  \sbox{\smOldLabel}{#1}\leavevmode
  \rlap{\hbox to\wd\smOldLabel{\hss\smash{\smFigureFace #2}\hss}}%
  \phantom{\usebox{\smOldLabel}}\endgroup}
\newcommand{\smLiteral}[1]{\smSameLabel{#1}{#1}}
\newcommand{\smResolution}[1]{\smSameLabel{\(#1\)}{{\def\times{\kern.15em\char128\kern.15em}#1}}}
\newcommand{\smMask}[4]{\put(#1,#2){\color{white}\rule{#3\unitlength}{#4\unitlength}}}
\newcommand{\smVideoRow}[4]{\put(#1,#2){\makebox(0,0){%
  \if\relax\detokenize{#4}\relax{\smFigureFace #3}\else
  \shortstack[c]{{\smFigureFace #3}\\{\smFigureSubFace #4}}\fi}}}

\newcommand{\paneltext}[3]{\put(#1,#2){\makebox(0,0){#3}}}
\newcommand{\resfigure}[6]{%
  \begingroup\normalfont\normalsize
  \sbox{\panelart}{\includegraphics[width=\linewidth]{HHM2K_#1_multires_uniform_art.pdf}}%
  \setlength{\unitlength}{\dimexpr\wd\panelart/510\relax}%
  \begin{picture}(510,145)
    \put(0,0){\usebox{\panelart}}
    \paneltext{50}{137}{\smLiteral{Input}}
    \paneltext{148.4}{137}{\smResolution{#2}}
    \paneltext{228}{137}{\smResolution{#3}}
    \paneltext{307.6}{137}{\smResolution{#4}}
    \paneltext{387.2}{137}{\smResolution{#5}}
    \paneltext{466.8}{137}{\smResolution{#6}}
    \paneltext{102.5}{100.5}{\rotatebox{90}{\smLiteral{Coarse}}}
    \paneltext{102.5}{33.8}{\rotatebox{90}{\textcolor[RGB]{0,94,90}{\smSameLabel{\bfseries Refined}{Refined}}}}
  \end{picture}\endgroup}

\newcommand{\reviewfigure}{%
  \begingroup\normalfont\normalsize
  \sbox{\panelart}{\includegraphics[width=\linewidth]{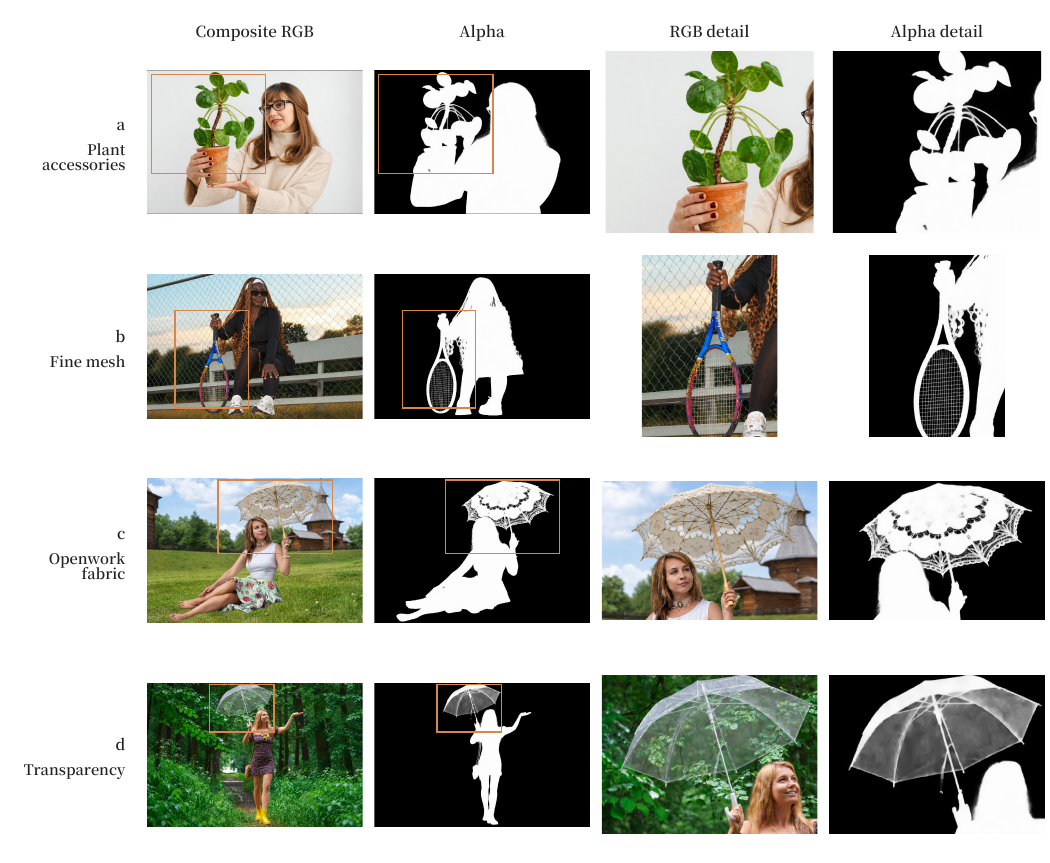}}%
  \setlength{\unitlength}{\dimexpr\wd\panelart/504\relax}%
  \begin{picture}(504,414)
    \put(0,0){\usebox{\panelart}}
  \end{picture}\endgroup}

\newcolumntype{Z}{>{\centering\arraybackslash}X}
\newcolumntype{J}{>{\raggedleft\arraybackslash}X}
\DeclareCaptionType{algorithm}[Algorithm][List of Algorithms]

\newlength{\smHalfWidth}

\definecolor{cvprblue}{rgb}{0.21,0.49,0.74}
\usepackage[pagebackref,breaklinks,colorlinks,allcolors=cvprblue]{hyperref}
\def\confName{CVPR}
\def\confYear{2027}
\title{SnapMat: Native-Resolution Video Matting\\via Context-Guided Sparse Refinement}
\author{%
  \normalsize Shijun Li\textsuperscript{*}\quad Yixin Yang\textsuperscript{*}\quad Yiheng Lin\textsuperscript{*}\quad
  Siyu Jiao\quad Zixin Ma\quad Yao Zhao\quad Yunchao Wei\textsuperscript{\textdagger}\\[3pt]
  Beijing Jiaotong University\\[2pt]
  {\small \textsuperscript{*}Equal contribution.\quad \textsuperscript{\textdagger}Corresponding author.}\\[2pt]
  {\normalsize\url{https://snapmat.github.io/}}}
\hypersetup{pdfauthor={Shijun Li, Yixin Yang, Yiheng Lin, Siyu Jiao, Zixin Ma, Yao Zhao, Yunchao Wei},pdftitle={SnapMat: Native-Resolution Video Matting via Context-Guided Sparse Refinement}}

\makeatletter
\patchcmd{\@maketitle}{\vspace*{24pt}}{\vspace*{5pt}}{}{\PackageError{snapmat}{Title spacing not found}{}}
\patchcmd{\@maketitle}{\vspace*{12pt}}{\vspace*{3pt}}{}{\PackageError{snapmat}{Author spacing not found}{}}
\g@addto@macro\@maketitle{%
  \noindent\begin{minipage}{\textwidth}
      \centering
      \includegraphics[width=0.84548\linewidth]{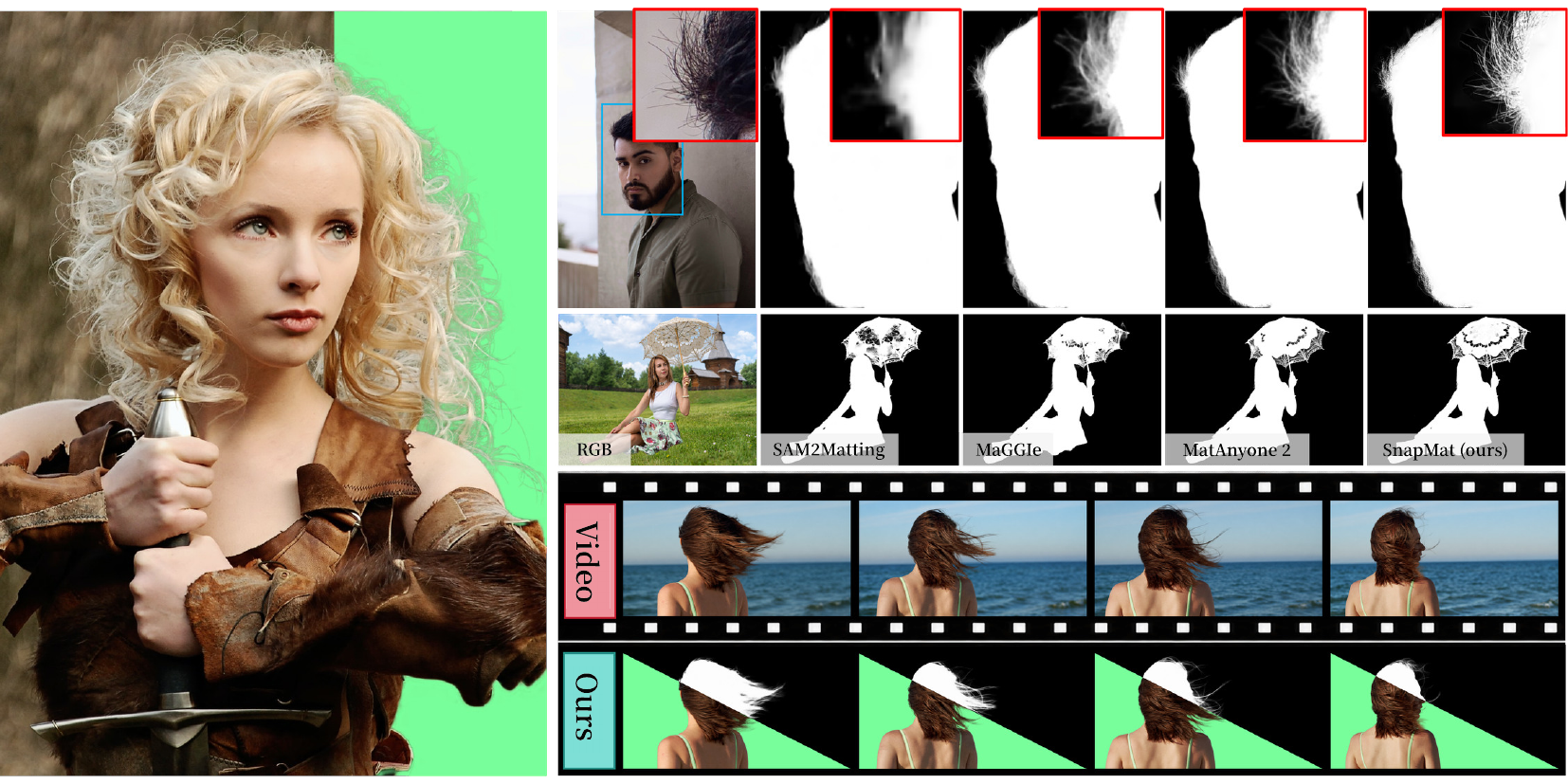}
      \captionsetup{hypcap=false}
      \captionof{figure}{SnapMat recovers fine structures at native resolution. We show image matting and compositing (left), comparisons with existing methods (top right), and video matting results across multiple frames (bottom right).}
      \label{fig:teaser}
  \end{minipage}\par\vspace{8pt}
}
\AtBeginDocument{\flushbottom}
\makeatother

\newif\ifincludelatersections
\includelatersectionstrue
\begin{document}
\maketitle
\patchcmd{\abstract}{\vspace*{12pt}}{\vspace*{2pt}}{}{}
\begin{abstract}
\tolerance=2300\looseness=-1
\spaceskip=.23em plus .10em minus .08em\relax
High-resolution video matting requires fine-detail recovery and temporal consistency, yet faces high computational costs and limited supervision. We propose Sparse Native-resolution Alpha Prediction, a sparse refinement framework that separates target propagation from detail recovery. Given a first-frame target mask, we adapt SAM2 with continuous alpha supervision for temporal propagation and coarse alpha prediction on the backbone grid. Dense Image-Context Encoding extracts low-resolution target and temporal context, which Sparse High-resolution Alpha Residual Prediction fuses with native-resolution RGB to predict alpha residuals in selected regions through two-dimensional sparse computation. Real-scene references guide coherent foreground--background generation, while generated RGBA layers provide foreground color and alpha. Compositing, review, and targeted correction yield the Reference-guided Alpha Layer Matting dataset, with 10,054 image samples and 1,007 four-frame clips. Our method achieves leading results across multiple evaluation settings on YouTubeMatte, VideoMatte240K, and HHM2K, with 4K inference at 21.76 FPS and 2.11 GB of peak allocated GPU memory on an NVIDIA RTX 5090. Ablations and equal-source-frame comparisons validate the refinement design and the training value of generated images and four-frame clips.
\par
\end{abstract}

\vspace{-16pt}%
\section{Introduction}\label{sec:intro}

High-resolution video matting aims to estimate continuous foreground opacity while preserving target completeness, fine structures, and temporal consistency. Reliable supervision must accurately capture hair strands, openwork structures, and semitransparent regions while maintaining consistent target definitions across frames. The high cost of such alpha annotations limits the scale of high-quality video matting datasets. When matting supervision is limited, pretrained target localization and temporal propagation provide a foundation for continuous alpha estimation without having to relearn these capabilities from scarce matting data.

\begingroup\tolerance=1800\looseness=-1
SAM2 \cite{ref11} acquires strong target localization and temporal propagation capabilities through large-scale pretraining. Recent work transfers these capabilities to matting through dedicated prediction components \cite{ref15,ref14} or direct fine-tuning with continuous alpha supervision \cite{ref18}. However, existing methods still face high computational costs \cite{ref14} or limitations in native-resolution detail recovery \cite{ref18,ref15}. High-resolution matting therefore requires an efficient way to combine pretrained target and temporal context with native image details at low computational cost. Dense feature computation over an entire frame at native resolution incurs substantial computational and GPU memory costs \cite{ref13}. Downsampling reduces these costs but may discard fine structures that upsampling cannot reliably recover. Existing sparse matting methods \cite{ref37,ref17,ref16} show that selectively allocating high-resolution computation to selected image regions can balance detail recovery and computational efficiency.\par\endgroup

\begingroup\tolerance=1800\looseness=-1
Unambiguous foreground interiors and background regions generally require little alpha correction, whereas boundaries, fine structures, and transition regions require more refinement. Target-level information generally varies less between frames, while fine details often change from frame to frame. This motivates assigning temporal propagation to the backbone and spatial detail recovery to a refinement branch. We therefore propose Sparse Native-resolution Alpha Prediction (SnapMat), which preserves low-resolution target and temporal context while concentrating native-resolution computation on selected regions. Given a first-frame target mask, we adapt SAM2 with continuous alpha supervision to predict coarse alpha and propagate target information. Dense Image-Context Encoding (DICE) provides low-resolution context; Sparse High-resolution Alpha Residual Prediction (SHARP) fuses it with native-resolution RGB to predict sparse alpha residuals. Temporal state remains on the backbone grid, and high-resolution feature computation is restricted to selected regions and their neighborhoods.\par\endgroup

\begingroup\tolerance=1800\looseness=-1
Fine-grained matting depends not only on model design but also on high-quality training supervision. Prior work expands matting data through generative priors \cite{ref18}, quality assessment \cite{ref09}, and layer generation \cite{ref35,ref34}. Generative models such as GPT Image 2.5 can directly produce high-quality RGBA images, providing foreground color and alpha in a single layer and enabling paired matting supervision. Building on this capability, we use real images and videos as references and specify which foreground entities to retain when generating foreground layers and backgrounds. The references guide scene layout, lighting, and target relationships so that the generated foregrounds and backgrounds remain visually natural and coherent. Compositing foreground color and alpha from the same RGBA layer yields pixel-aligned RGB--alpha training pairs, while review and targeted correction address generation errors. We extend the pipeline to four-frame video clips with joint review for temporal consistency. The resulting Reference-guided Alpha Layer Matting (REALM) dataset contains 10,054 image samples and 1,007 four-frame clips, covering fine structures, semitransparent materials, and scenes in which people and designated associated objects jointly form the foreground.\par\endgroup

\begingroup\tolerance=1800\looseness=-1
Experiments show that SnapMat achieves leading results across multiple evaluation settings on YouTubeMatte, VideoMatte240K, and HHM2K. Module and resolution ablations support the benefits of context fusion and native-resolution refinement, respectively. On an NVIDIA RTX 5090, SnapMat achieves 21.76 FPS at 4K with 2.11 GB of peak allocated GPU memory. Full-model data ablations validate the training value of REALM. Controlled comparisons with equal source-frame counts further support the effectiveness of four-frame clips for adapting the pretrained backbone under the current training protocol. Our two main contributions are summarized below:\par\endgroup
\begin{enumerate}
\item \textbf{Context-guided native-resolution sparse refinement.} We use SAM2 target and temporal context to guide a two-dimensional sparse branch that predicts native-resolution alpha residuals for efficient detail recovery.
\item \textbf{Real-reference-guided matting data construction.} We leverage high-quality RGBA generation to obtain foreground color and alpha, while real-scene references guide coherent foreground--background generation. Compositing, review, and correction yield REALM images and four-frame clips, evaluated through data ablations and matched-source-frame comparisons.
\end{enumerate}

\section{Related Work}\label{sec:related}
\begin{figure*}[!t]
  \centering
  \includegraphics[width=0.84\textwidth]{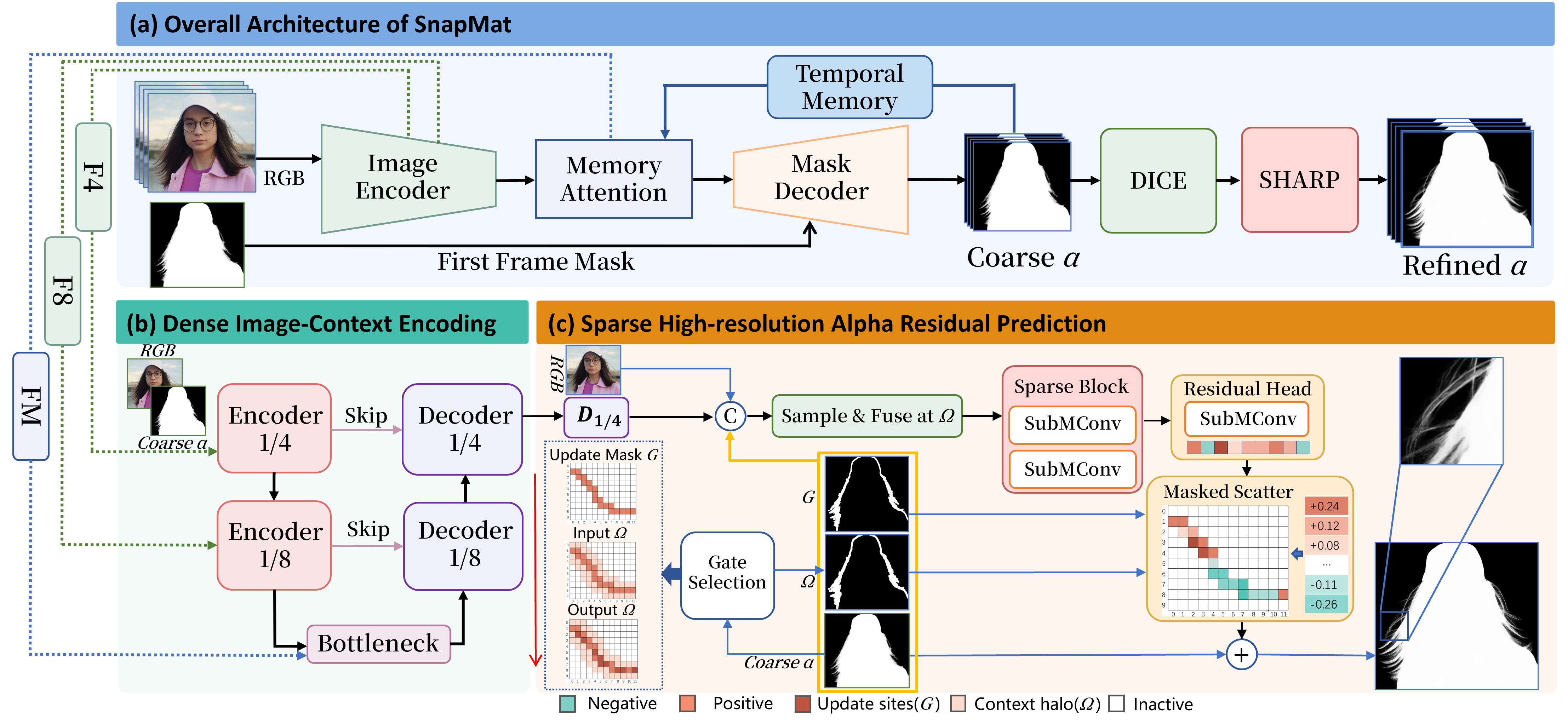}
  \caption{Overview of SnapMat. SAM2 provides coarse alpha and temporal context. DICE fuses low-resolution context, and SHARP predicts native-resolution residuals at sparse locations selected from the coarse alpha.}
  \label{fig:architecture}
\end{figure*}

\subsection{Image and Video Matting}

\begingroup\tolerance=3000\looseness=-1
Matting first requires defining the foreground target. Conventional methods rely on trimaps, clean backgrounds, or coarse masks \cite{ref50,ref04,ref01,ref49,ref03}, automatic portrait matting uses category priors \cite{ref37,ref02,ref16}, and Click2Trimap \cite{zhang2025learning} converts user clicks into trimaps. In video matting, target information must also propagate across frames to handle motion and occlusion while maintaining consistent predictions. To address this temporal requirement, RVM \cite{ref05} uses recurrent mechanisms, MatAnyone \cite{ref08} employs memory propagation, and MatAnyone 2 \cite{ref09} further incorporates quality assessment and temporally distant reference frames to improve training supervision and appearance modeling. Beyond designing temporal architectures specifically for matting, another approach transfers the pretrained capabilities of segmentation models.\par\endgroup

\begingroup\tolerance=1800\looseness=-1
In segmentation, SAM \cite{ref10} supports prompt-driven target selection, which SAM2 \cite{ref11} extends to propagation in videos. To move from binary segmentation to continuous alpha estimation, MAM2 \cite{ref14} augments SAM2 with dual-branch decoding for masks and trimaps, decoupled memory, and a separate matting module. SAM2Matting \cite{ref15} freezes the segmentation backbone and connects tracking with matting through region detection and progressive alpha prediction. SAM2-Matte in VideoMaMa \cite{ref18} directly adapts SAM2 using continuous alpha supervision. Beyond segmentation pretraining, VideoMaMa++ \cite{ref53} uses video diffusion priors to improve target structure and temporal consistency.\par\endgroup

\begingroup\tolerance=1800\looseness=-1
These methods improve target and alpha estimation, but recovering high-resolution details also requires controlling computational and GPU memory costs. MEMatte \cite{lin2025memory} reduces memory consumption through adaptive token routing, while MatAnyone-E \cite{ref52} uses sparse memory propagation and a lightweight decoder. For refinement, SparseMat \cite{ref16} selects regions based on low-resolution predictions and interframe differences, then recovers alpha with a sparse encoder and decoder. $\mu$Matting \cite{ref37} uses three-dimensional sparse convolutions and temporal context modulation, while MaGGIe \cite{ref17} combines instance-level sparse features with multiscale dense features to progressively predict and \mbox{fuse alpha estimates}. These approaches control computation through memory propagation, decoding, and local refinement. We use backbone context for sparse alpha correction at native resolution (Section~\ref{sec:method}).\par\endgroup

\subsection{Matting Datasets}

\begingroup\tolerance=1800\looseness=-1
Matting performance also depends on high-quality training supervision. Existing resources include portrait data represented by P3M-10K \cite{ref28,ref41,sun2022instmatt,ref16}, multicategory foreground data represented by Composition-1k \cite{ref27,ref01}, animal data AM-2K \cite{ref29}, transparent-object data Transparent-460 \cite{ref42}, and video data represented by VideoMatte240K \cite{ref04,ref23,ref43}. The annotation costs of fine structures, semitransparent regions, and interframe consistency constrain data expansion. Many datasets, including VM800 \cite{ref08} and VMReal \cite{ref09}, remain unavailable for various reasons, further limiting access to video supervision. To expand supervision, prior work automatically constructs or filters annotations from existing material. HHM50K \cite{ref16} expands high-resolution portrait data, and COCO-Matting \cite{ref13} converts segmentation masks to alpha after determining foreground membership for associated objects. VideoMaMa \cite{ref18} uses generative priors to construct video pseudo-labels, while MatAnyone 2 \cite{ref09} identifies reliable annotations through \mbox{pixel-level quality assessment}.\par\endgroup

\begingroup\tolerance=3000\looseness=-1
Beyond annotating existing material, generation and rendering provide new content. Layer Diffusion \cite{ref34}, LD-Portrait-20K \cite{ref35}, GVM \cite{ref19}, and related work \cite{ref44,ref45,ref47,ref48,ref46} explore the generation and rendering of transparent content and matting data. However, both automatic annotation and data generation require inspection and correction of target omissions, boundary errors, and residual backgrounds. We leverage generative models to obtain foreground color and alpha from high-quality RGBA layers, while real-scene references guide appearance and foreground--background relationships. Compositing and targeted correction yield REALM image and video data. Section~\ref{sec:data} details the pipeline.\par\endgroup

\section{Method}\label{sec:method}

\subsection{Overview}

\begingroup\tolerance=3000\looseness=-1
As shown in Figure~\ref{fig:architecture}, SnapMat separates target propagation from detail recovery. The adapted SAM2 performs target localization, temporal propagation, and coarse alpha prediction on the backbone grid. DICE extracts low-resolution target context, and SHARP fuses it with native-resolution RGB to predict alpha residuals at selected locations, recovering fine details using propagated target and temporal context.\par\endgroup

\subsection{Adapting SAM2 to Video Matting}

We retain the main architecture of SAM2 and adapt it to opacity estimation through continuous alpha supervision. A binary mask in the first frame specifies the target, and the memory mechanism propagates target information to subsequent frames. Let $Z_t$ denote the current-frame logits produced by the SAM2 mask decoder. The coarse alpha used by the refinement branch is defined as
\begin{equation}
\alpha_t^c=\mathcal R_t\bigl(\sigma(Z_t)\bigr).
\end{equation}
\begingroup\tolerance=1800\looseness=-1
Here, $\sigma$ is the sigmoid function, and $\mathcal R_t$ resizes the coarse alpha to match the spatial resolution of the RGB input used for refinement. Memory updates still follow the coarse prediction path. Refined predictions are not directly written back to memory. Since restoring the prediction size cannot recover lost fine structures, we further use coarse alpha to select regions for native-resolution correction.\par\endgroup

\subsection{Coarse Alpha-Guided Sparse Refinement}

To select regions for sparse correction, we first combine coarse alpha transitions with the binary target contour, then dilate their union to obtain $G$:
\begin{equation}
\begin{aligned}
U&=\mathbf1[0.05<\alpha^c<0.95],\\
G&=\mathcal D_r\bigl(U\cup\partial\mathbf1[\alpha^c\geq0.5]\bigr).
\end{aligned}
\end{equation}
Here, $\partial$ extracts the binary contour, and $\mathcal D_r$ denotes dilation with radius $r$. Including the contour covers nearly binary boundaries that still require correction. To incorporate nearby foreground and background information, we construct the sparse computation set used by SHARP:
\begin{equation}
\Omega=\operatorname{supp}\bigl(\mathcal D_h(G)\bigr).
\end{equation}
Here, $h$ denotes the extra dilation radius, and $\operatorname{supp}$ selects nonzero locations. While $G$ marks where alpha is updated, $\Omega$ also includes neighboring pixels required for sparse convolution without changing the update region.

\begingroup\tolerance=1800\looseness=-1
The selected regions determine where to compute, but interpreting local RGB also requires target and historical information. DICE therefore first extracts low-resolution context: it encodes RGB and coarse alpha, fuses multiscale SAM2 image features, and incorporates historical memory features at the bottleneck. Upsampling and skip connections produce $D_{1/4}$ at one-quarter of the refinement RGB input resolution in each spatial dimension.\par\endgroup

\begingroup\tolerance=1800\looseness=-1
This context identifies the target, while native-resolution RGB supplies local textures and boundaries. SHARP maps coordinates in $\Omega$ to $D_{1/4}$, samples and projects the context features, and concatenates them with RGB, coarse alpha, and the region indicator at corresponding locations. It aggregates neighborhood information through $3\times3$ submanifold sparse convolutions (SubMConv) \cite{ref21}, normalization, and nonlinear transformations. SubMConv preserves the active location set. The branch performs only two-dimensional spatial computation. Temporal information comes from the backbone context.\par\endgroup

The residual head converts the fused features into corrections $\Delta\alpha$, scatters them back onto the image grid, and updates alpha only within $G$:
\begin{equation}
\alpha^r=\operatorname{clip}\bigl(\alpha^c+G\odot\Delta\alpha,0,1\bigr).
\end{equation}
Here, $\odot$ denotes elementwise multiplication, and the coarse prediction is preserved outside the region. Figure~\ref{fig:coarse-refined} compares local details in the coarse and refined predictions.

\begingroup
\setlength{\intextsep}{2.5pt plus 1pt minus 0.5pt}
\begin{figure}[H]
  \centering
  \includegraphics[width=0.84\linewidth]{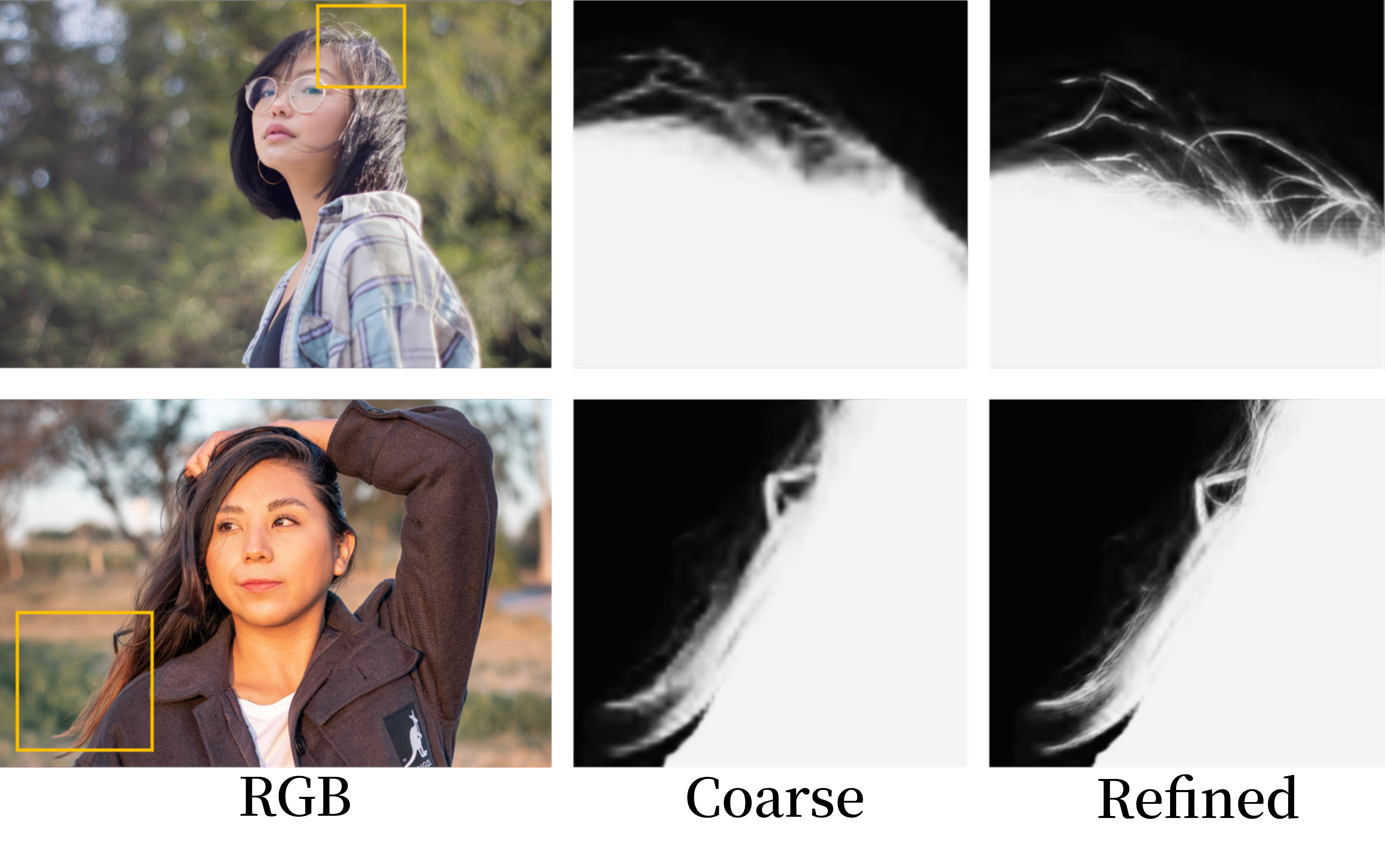}
  \caption{Comparison of coarse and refined alpha mattes.}
  \label{fig:coarse-refined}
\end{figure}
\endgroup

\begin{figure*}[!t]
  \centering
  \includegraphics[width=0.90\textwidth]{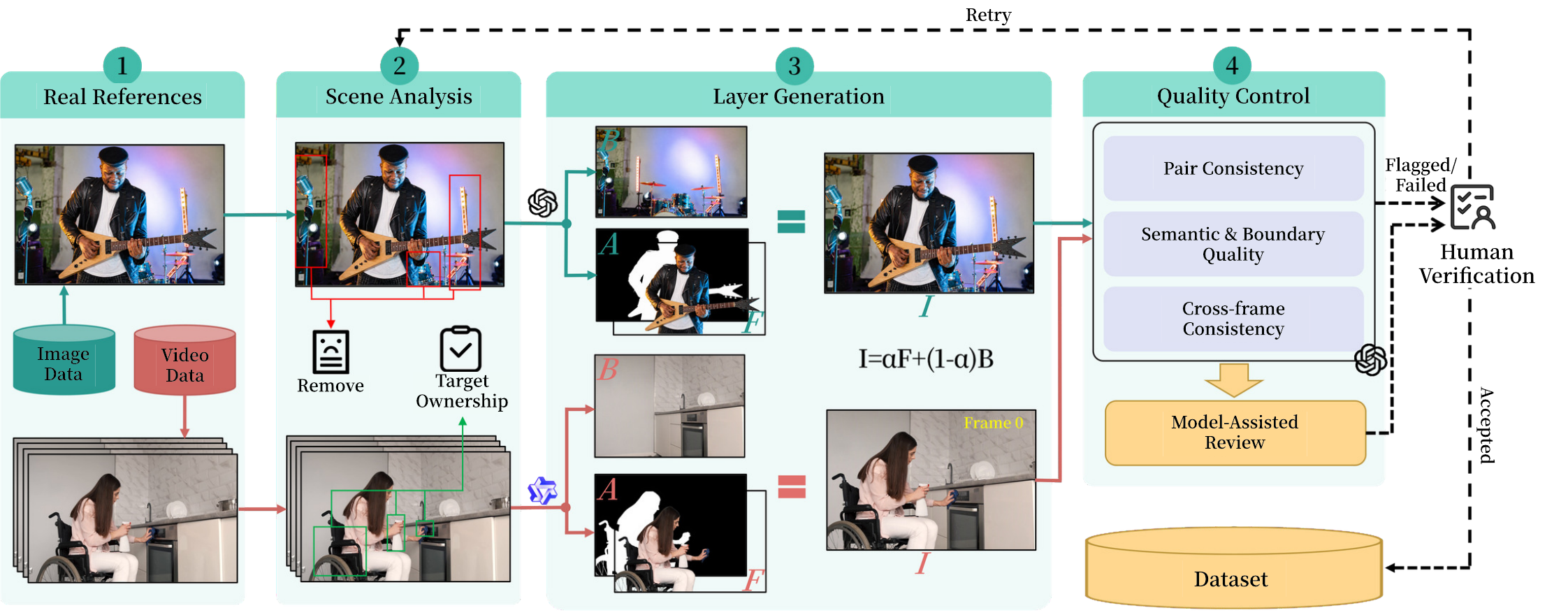}
  \caption{REALM data construction. Real references and target constraints guide layer generation, which is followed by compositing, quality review, and targeted correction of generated layers before final acceptance.}
  \label{fig:data-pipeline}
\end{figure*}

\subsection{Staged Optimization}\label{subsec:staged-optimization}

We train SnapMat in three progressive stages: adapting SAM2 for coarse alpha prediction, training the refinement modules with the adapted backbone frozen, and jointly optimizing coarse prediction and refinement. All three stages use pixelwise absolute error, Laplacian pyramid error, and gradient error, weighted by 5, 2, and 1, respectively. Let $\alpha^*$ denote ground-truth alpha and $M$ the supervision~region.~Then
\begin{equation}
\mathcal L_{\mathrm{mat}}(a,\alpha^*;M)
=5\mathcal L_1+2\mathcal L_{\mathrm{lap}}+\mathcal L_{\mathrm{grad}}.
\end{equation}

\noindent\textbf{Backbone adaptation.} Starting from publicly available pretrained weights, we first fine-tune all SAM2 parameters to learn continuous coarse alpha prediction. To strengthen pixelwise supervision in the alpha transition region, we compute mean absolute error separately within the ground-truth transition region and its complement, and sum the two terms.

\noindent\textbf{Refinement module training.} We then freeze the adapted SAM2 and train DICE and SHARP using the coarse predictions and context features produced by \mbox{the frozen backbone.} The refined alpha is supervised within the selected region by $\mathcal L_{\mathrm{mat}}(\alpha_t^r,\alpha_t^*;G_t)$. The residual head is zero-initialized so that the initial refined prediction \mbox{matches the coarse prediction.}

\noindent\textbf{Joint optimization.} Once refinement has learned to use backbone outputs, joint optimization coordinates backbone features and residual prediction to improve coarse alpha and local corrections together. Starting from the trained components, we jointly update the image encoder, feature fusion neck, mask decoder, DICE, and SHARP, while keeping prompt- and memory-related modules frozen. Coarse and refined predictions receive equally weighted supervision over the valid region $V_t$ and refinement region $G_t$, respectively. Letting \mbox{$\ell_t(a;M)=\mathcal L_{\mathrm{mat}}(a,\alpha_t^*;M)$, we obtain}
\begin{equation}
\mathcal L_{\mathrm{joint}}
=\sum_t\bigl[\ell_t(\alpha_t^c;V_t)+\ell_t(\alpha_t^r;G_t)\bigr].
\end{equation}
Refinement training and joint optimization both average pixelwise errors over the corresponding supervision regions. Region and coordinate selection are excluded from gradient computation. Complete loss definitions and implementation details are provided in the appendix.

\section{Reference-Guided Matting Data Generation}\label{sec:data}

\begingroup\tolerance=1800\looseness=-1
Generative models such as GPT Image 2.5 can directly produce high-quality RGBA images containing foreground color and alpha, providing the components needed for paired matting supervision. To maintain realistic scene appearance and target relationships, we use real images and videos as references and explicitly define which foreground entities to retain when guiding foreground-layer and background generation. We composite foreground color and alpha from the same RGBA layer with generated backgrounds, then review and correct the resulting samples. Figure~\ref{fig:data-pipeline} illustrates the image and video data construction pipelines.\par\endgroup

\subsection{Reference Selection and Target Definition}

\begingroup\tolerance=1800\looseness=-1
We select real images and videos spanning diverse scenes, materials, and person--object interactions. Since one scene can have different foreground definitions, we specify which people, worn items, and associated objects to retain. This definition guides generation, review, and training prompt construction. Reference sources, licenses, and detailed target rules are provided in the appendix.\par\endgroup

\subsection{Image and Video Supervision Construction}

\begingroup\tolerance=1800\looseness=-1
\noindent\textbf{Layer generation.} Given real references and a foreground definition, the foreground branch uses an RGBA generation model to produce foreground color and alpha for the designated targets, while the background branch removes these targets and completes the scene. Shared scene references guide coherent layout, lighting, and appearance. Generation instructions specify opacity requirements for opaque entities, holes, and semitransparent materials, guiding alpha to match foreground structures.\par\endgroup

\begingroup\tolerance=1800\looseness=-1
\noindent\textbf{Compositing and supervision pairing.} We obtain foreground color and alpha from the same RGBA layer and composite them with the generated background to form supervision. With non-premultiplied foreground color $F$, opacity $\alpha$, and background $B$, the training image is
\begin{equation}
I=\alpha\odot F+(1-\alpha)\odot B.
\end{equation}
Here, $\odot$ denotes elementwise multiplication. The resulting $(I,\alpha)$ pairs are pixel-aligned.\par\endgroup

\begin{figure*}[!t]
  \centering
  \includegraphics[width=0.83\textwidth,trim=0 4bp 0 0,clip]{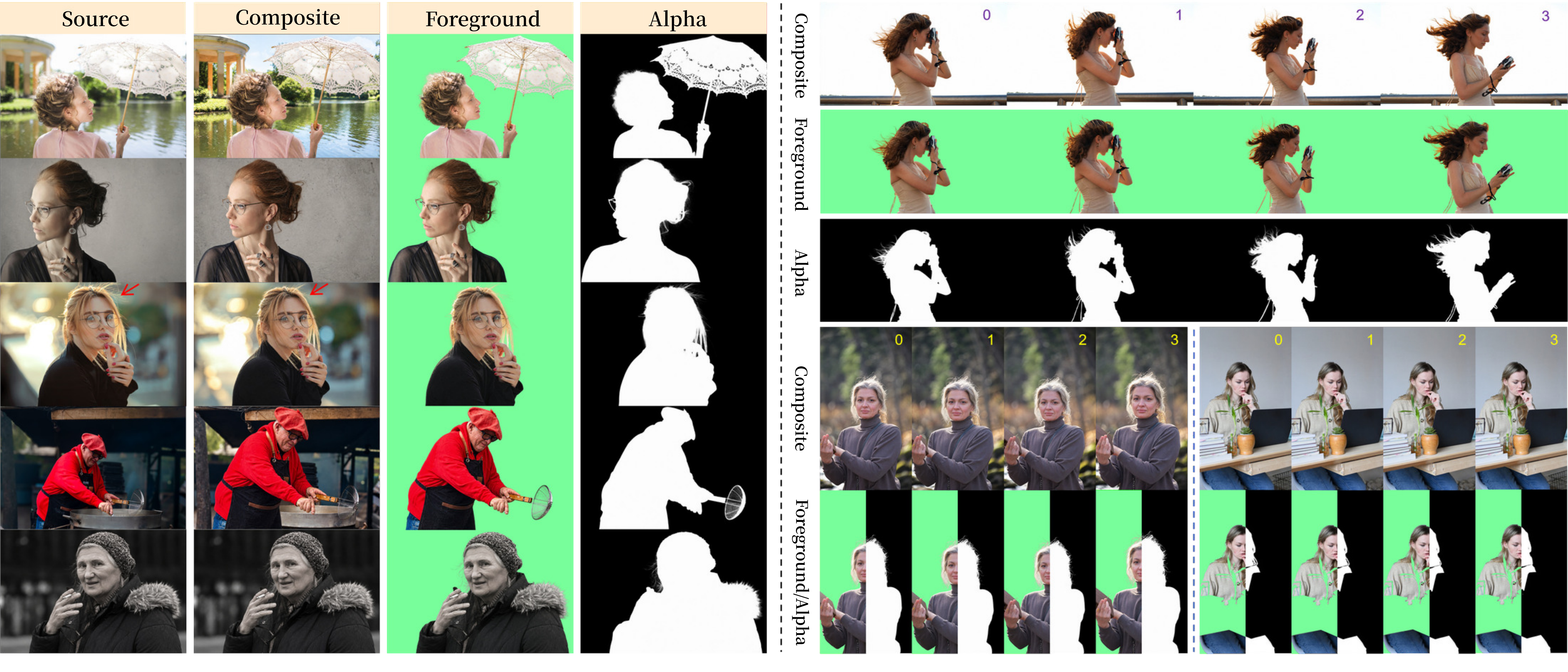}
  \caption{REALM image samples (left) and four-frame video clips (right), together with their generated layers.}
  \label{fig:realm-samples}
\end{figure*}

\begingroup\tolerance=1800\looseness=-1
\noindent\textbf{Four-frame video units.} Video pairs require consistent foreground membership, structure, and transparency across frames. We sample four frames from one shot at approximately 0.2-second intervals and apply the same generation and compositing process to each. They share scene analysis and target definitions, but layers are generated independently. Each complete clip therefore undergoes joint review, confining temporal checks and corrections to a short window that matches frame-pair sampling during training.\par\endgroup

\begingroup\tolerance=1800\looseness=-1
\noindent\textbf{Quality review and layer correction.} Pixel alignment does not rule out missing targets, residual backgrounds, or implausible transparency. Programmatic checks verify layer pairing and compositing. A vision model assesses target completeness and the plausibility of transparency. Complex materials and ambiguous samples undergo human review. We regenerate only affected layers and recheck the samples. Corrected videos are accepted only after the entire clip passes another joint review. Detailed checks, cleanup rules, generation and review settings, and revision records are provided in the appendix.\par\endgroup

\begingroup\tolerance=1800\looseness=-1
The resulting REALM dataset contains 10,054 image samples in REALM-Image10K and 1,007 four-frame clips in REALM-Video1K, covering fine structures, semitransparent materials, and foregrounds comprising people and designated associated objects. Figure~\ref{fig:realm-samples} presents image samples, four-frame clips, and their corresponding generated foreground and alpha layers.\par\endgroup

\begingroup

\newdimen\NBbenchmarkheight
\newdimen\NBbenchmarkrowspace
\newsavebox{\NBbenchmarkbodybox}
\newsavebox{\NBbenchmarkcaptionbox}
\NBbenchmarkheight=431.5pt
\setlength{\intextsep}{0pt}
\nointerlineskip
\begin{table}[!t]
\centering
\renewcommand{\NBmetric}[1]{#1$\downarrow$}
\renewcommand{\NBplain}[1]{#1}
\renewcommand{\NBbest}[1]{\textbf{#1}}
\renewcommand{\NBsecond}[1]{#1}
\fontsize{8.2}{7.8}\selectfont
\setlength{\tabcolsep}{1.5pt}
\setlength{\arrayrulewidth}{0.3pt}
\setlength{\aboverulesep}{0.85pt}
\setlength{\belowrulesep}{0.85pt}
\renewcommand{\arraystretch}{1.06}
\captionsetup{position=top,skip=4pt}
\NBbenchmarkrowspace=0pt
\newcommand{\NBbenchmarkrowgap}{\noalign{\vskip\NBbenchmarkrowspace}}
\renewcommand{\NBdataset}[2]{\noalign{\vskip 2.0pt}\multicolumn{5}{@{}l@{}}{\textbf{\textit{#1}}\enspace #2}\\\NBbenchmarkrowgap}
\newcommand{\NBbenchmarkbody}{%
\begin{tabular*}{\linewidth}{@{\extracolsep{\fill}}l|cccc@{}}
\toprule
Metrics & SAM2Matting & MaGGIe + SAM2 & MatAnyone 2 & \textbf{SnapMat} \\\NBbenchmarkrowgap
\midrule
\NBdataset{YouTubeMatte}{($512\times288$)}
\midrule
\NBmetric{MAD}   & \NBplain{3.74} & \NBplain{3.46} & \NBsecond{2.30} & \NBbest{1.91} \\\NBbenchmarkrowgap
\NBmetric{MSE}   & \NBplain{1.81} & \NBplain{1.53} & \NBsecond{0.78} & \NBbest{0.53} \\\NBbenchmarkrowgap
\NBmetric{Grad}  & \NBplain{1.46} & \NBplain{0.88} & \NBsecond{0.78} & \NBbest{0.33} \\\NBbenchmarkrowgap
\NBmetric{Conn}  & \NBplain{0.54} & \NBplain{0.50} & \NBsecond{0.33} & \NBbest{0.22} \\\NBbenchmarkrowgap
\NBmetric{dtSSD} & \NBplain{2.01} & \NBplain{1.53} & \NBsecond{1.45} & \NBbest{1.00} \\\NBbenchmarkrowgap
\midrule
\NBdataset{YouTubeMatte}{($1920\times1080$)}
\midrule
\NBmetric{MAD}   & \NBplain{3.13} & \NBplain{2.73} & \NBsecond{1.68} & \NBbest{1.27} \\\NBbenchmarkrowgap
\NBmetric{MSE}   & \NBplain{1.50} & \NBplain{1.35} & \NBsecond{0.57} & \NBbest{0.28} \\\NBbenchmarkrowgap
\NBmetric{Grad}  & \NBplain{13.01} & \NBsecond{7.32} & \NBplain{7.46} & \NBbest{3.58} \\\NBbenchmarkrowgap
\NBmetric{Conn}  & \NBplain{6.35} & \NBplain{5.58} & \NBsecond{3.32} & \NBbest{2.22} \\\NBbenchmarkrowgap
\NBmetric{dtSSD} & \NBplain{2.11} & \NBplain{1.69} & \NBsecond{1.55} & \NBbest{1.22} \\\NBbenchmarkrowgap
\midrule
\NBdataset{VideoMatte240K}{($512\times288$)}
\midrule
\NBmetric{MAD}   & \NBplain{4.94} & \NBplain{4.83} & \NBsecond{4.73} & \NBbest{4.24} \\\NBbenchmarkrowgap
\NBmetric{MSE}   & \NBplain{0.53} & \NBsecond{0.36} & \NBplain{0.55} & \NBbest{0.19} \\\NBbenchmarkrowgap
\NBmetric{Grad}  & \NBplain{0.49} & \NBsecond{0.35} & \NBplain{0.50} & \NBbest{0.18} \\\NBbenchmarkrowgap
\NBmetric{Conn}  & \NBplain{0.24} & \NBplain{0.22} & \NBsecond{0.20} & \NBbest{0.11} \\\NBbenchmarkrowgap
\NBmetric{dtSSD} & \NBplain{1.52} & \NBsecond{1.08} & \NBplain{1.12} & \NBbest{0.80} \\\NBbenchmarkrowgap
\midrule
\NBdataset{VideoMatte240K}{($1920\times1080$)}
\midrule
\NBmetric{MAD}   & \NBplain{4.73} & \NBplain{4.29} & \NBbest{4.02} & \NBsecond{4.08} \\\NBbenchmarkrowgap
\NBmetric{MSE}   & \NBplain{0.50} & \NBplain{0.29} & \NBsecond{0.22} & \NBbest{0.21} \\\NBbenchmarkrowgap
\NBmetric{Grad}  & \NBplain{6.77} & \NBplain{3.71} & \NBsecond{2.96} & \NBbest{2.88} \\\NBbenchmarkrowgap
\NBmetric{Conn}  & \NBplain{3.12} & \NBplain{1.99} & \NBbest{1.47} & \NBsecond{1.52} \\\NBbenchmarkrowgap
\NBmetric{dtSSD} & \NBplain{1.74} & \NBplain{1.23} & \NBsecond{1.13} & \NBbest{1.11} \\\NBbenchmarkrowgap
\midrule
\NBdataset{VideoMatte240K}{($3840\times2160$)}
\midrule
\NBmetric{MAD}   & \NBplain{4.60} & \NBplain{4.25} & \NBplain{3.84} & \NBbest{3.81} \\\NBbenchmarkrowgap
\NBmetric{MSE}   & \NBplain{0.44} & \NBplain{0.29} & \NBplain{0.19} & \NBbest{0.15} \\\NBbenchmarkrowgap
\NBmetric{Grad}  & \NBplain{22.18} & \NBplain{14.13} & \NBbest{7.69} & \NBplain{8.05} \\\NBbenchmarkrowgap
\NBmetric{Conn}  & \NBplain{11.87} & \NBplain{8.31} & \NBplain{4.84} & \NBbest{4.49} \\\NBbenchmarkrowgap
\NBmetric{dtSSD} & \NBplain{1.77} & \NBplain{1.44} & \NBplain{1.07} & \NBbest{1.04} \\\NBbenchmarkrowgap
\midrule
\NBdataset{HHM2K}{(long side $\leq1920$)}
\midrule
\NBmetric{MAD}   & \NBsecond{4.32} & \NBplain{10.66} & \NBplain{8.87} & \NBbest{3.19} \\\NBbenchmarkrowgap
\NBmetric{MSE}   & \NBsecond{1.27} & \NBplain{6.76} & \NBplain{5.52} & \NBbest{0.79} \\\NBbenchmarkrowgap
\NBmetric{Grad}  & \NBplain{34.18} & \NBplain{36.86} & \NBsecond{31.35} & \NBbest{22.44} \\\NBbenchmarkrowgap
\NBmetric{Conn}  & \NBsecond{56.37} & \NBplain{148.55} & \NBplain{108.63} & \NBbest{39.25} \\\NBbenchmarkrowgap
\bottomrule
\end{tabular*}%
}
\sbox{\NBbenchmarkbodybox}{\NBbenchmarkbody}
\setbox\NBbenchmarkcaptionbox=\vbox{%
\hsize=\linewidth
\caption{Quantitative comparison. Best values before rounding are in bold. All HHM2K inputs have a 1920-pixel long-side cap, with outputs upsampled to the original image size for scoring. The VideoMatte240K ($3840\times2160$) block reports VM4K results; construction and inference settings are described in Section~5.2.}
\label{tab:benchmark}
}
\NBbenchmarkrowspace=\dimexpr\NBbenchmarkheight-\ht\NBbenchmarkbodybox-\dp\NBbenchmarkbodybox-\ht\NBbenchmarkcaptionbox-\dp\NBbenchmarkcaptionbox\relax
\divide\NBbenchmarkrowspace by 36
\ifdim\NBbenchmarkrowspace<0pt \NBbenchmarkrowspace=0pt\fi
\box\NBbenchmarkcaptionbox
\nointerlineskip
\NBbenchmarkbody\par
\end{table}
\endgroup

\flushbottom
\ifincludelatersections
  \begingroup
  \clubpenalty=10000
  \widowpenalty=10000
  \brokenpenalty=100
  \vspace{-2.5pt}%
\section{Experiments}\label{sec:experiments}

\vspace{-3pt}%
\subsection{Training Setup}\label{subsec:training}

\begingroup\tolerance=1800\looseness=-1
We initialize from publicly available SAM2.1 Hiera-L pretrained weights. Existing training data include videos from VideoMatte240K \cite{ref04} and CRGNN \cite{ref23}, and images from P3M-10K \cite{ref28}, AM-2K \cite{ref29}, and Human-2K \cite{ref41}, among others. Backbone adaptation uses existing multi-source data, while refinement module training and joint optimization additionally mix REALM with existing images and videos. Each sample contains two frames, with a target mask provided only for the first frame. SAM2 takes $1024\times1024$ inputs, and the refinement branch uses crops with varying aspect ratios. Subsequent ablations use this configuration as their baseline. Detailed training settings, update counts, and data compositions are provided in the appendix.\par\endgroup

\begin{figure*}[!t]
  \centering
  \includegraphics[width=0.82\textwidth,trim=0 5bp 0 0,clip]{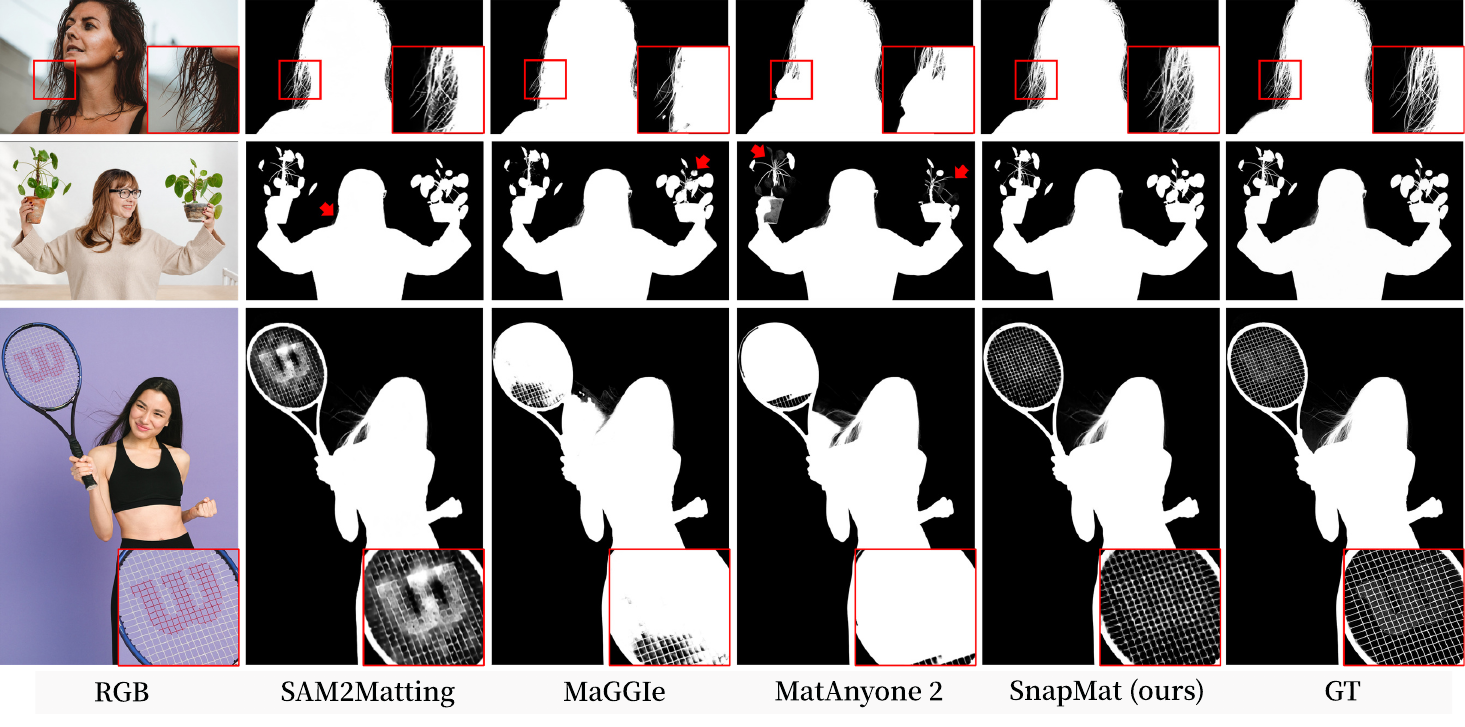}
  \caption{Qualitative comparison of fine structures and target completeness. Insets and arrows highlight details.}
  \label{fig:qualitative-comparison}
\end{figure*}

\begin{table*}[!b]
\centering
\caption{Inference efficiency and memory usage. Gray values denote SAM2Matting's fixed $1008\times1008$ grid with output resizing, not native-resolution inference. OOM denotes insufficient GPU memory at that resolution.}
\label{tab:efficiency}
\fontsize{8.2}{9.6}\selectfont
\setlength{\tabcolsep}{3pt}
\renewcommand{\arraystretch}{1.0}
\newcommand{\NBspeedcell}[2]{#1 & / & #2}
\newcommand{\NBgrayspeedcell}[2]{\textcolor{black!20}{#1} & \textcolor{black!20}{/} & \textcolor{black!20}{#2}}
\begin{tabular*}{\textwidth}{@{\extracolsep{\fill}}ll*{4}{>{\centering\arraybackslash}p{22pt}@{\extracolsep{0pt}\hspace{4pt}}>{\centering\arraybackslash}p{6pt}@{\hspace{4pt}}>{\centering\arraybackslash}p{22pt}@{\extracolsep{\fill}}}@{}}
\toprule
 & & \multicolumn{3}{c}{$1024\times1024$} & \multicolumn{3}{c}{$1920\times1080$} & \multicolumn{3}{c}{$2560\times1440$} & \multicolumn{3}{c}{$3840\times2160$} \\
\textbf{GPU} & \textbf{Method} & \NBspeedcell{FPS$\uparrow$}{GB$\downarrow$} & \NBspeedcell{FPS$\uparrow$}{GB$\downarrow$} & \NBspeedcell{FPS$\uparrow$}{GB$\downarrow$} & \NBspeedcell{FPS$\uparrow$}{GB$\downarrow$} \\
\midrule
\textbf{RTX 4090} & SnapMat & \NBspeedcell{28.43}{1.44} & \NBspeedcell{18.87}{1.78} & \NBspeedcell{18.22}{1.84} & \NBspeedcell{16.57}{2.11} \\
 & SAM2Matting & \NBspeedcell{12.91}{2.93} & \NBgrayspeedcell{12.92}{2.94} & \NBgrayspeedcell{12.90}{2.96} & \NBgrayspeedcell{12.91}{3.00} \\
 & MatAnyone 2 & \NBspeedcell{26.28}{1.66} & \NBspeedcell{11.66}{5.82} & \NBspeedcell{4.74}{17.35} & \NBspeedcell{OOM}{OOM} \\
 & MaGGIe + SAM2 & \NBspeedcell{9.12}{3.03} & \NBspeedcell{5.90}{4.65} & \NBspeedcell{3.59}{7.29} & \NBspeedcell{1.69}{14.50} \\
\midrule
\textbf{RTX 5090} & SnapMat & \NBspeedcell{34.43}{1.44} & \NBspeedcell{24.16}{1.78} & \NBspeedcell{23.47}{1.84} & \NBspeedcell{21.76}{2.11} \\
 & SAM2Matting & \NBspeedcell{17.99}{2.93} & \NBgrayspeedcell{17.97}{2.94} & \NBgrayspeedcell{17.95}{2.96} & \NBgrayspeedcell{17.46}{3.00} \\
 & MatAnyone 2 & \NBspeedcell{43.00}{1.66} & \NBspeedcell{18.82}{5.82} & \NBspeedcell{7.83}{17.35} & \NBspeedcell{OOM}{OOM} \\
 & MaGGIe + SAM2 & \NBspeedcell{12.43}{3.03} & \NBspeedcell{7.86}{4.65} & \NBspeedcell{4.87}{7.29} & \NBspeedcell{2.40}{14.50} \\
\bottomrule
\end{tabular*}
\end{table*}

\vspace{-4pt}%
\subsection{Performance Comparison}

\begingroup\tolerance=1800\looseness=-1
We evaluate on VideoMatte240K (VM240K) \cite{ref04}, YouTubeMatte \cite{ref08}, and HHM2K \cite{ref16}. MatAnyone 2 \cite{ref09}, MaGGIe \cite{ref17}, SAM2Matting \cite{ref15}, and SnapMat use prompts derived from the same source, with all predictions resized to the annotation dimensions for scoring. We report mean absolute difference (MAD), mean squared error (MSE), gradient error (Grad), and connectivity error (Conn) \cite{ref51}, together with dtSSD for videos. Lower values indicate better performance across all reported metrics. The video benchmarks assess temporal consistency alongside alpha accuracy, while HHM2K evaluates fine structures in high-resolution images. \par\Needspace{5\baselineskip}\noindent SnapMat uses the same final checkpoint throughout, producing coarse alpha at $512\times288$ and enabling refinement at higher resolutions and on HHM2K. All HHM2K inputs have a 1920-pixel long-side limit. We construct a $3840\times2160$ test set, denoted VM4K, by compositing VideoMatte240K test foregrounds and their alpha mattes with BG20K backgrounds. MatAnyone 2 performs inference at $2560\times1440$ and upsamples its predictions to 4K for scoring.\par\endgroup

\begingroup\tolerance=1800\looseness=-1
As shown in Table~\ref{tab:benchmark}, SnapMat achieves the best results on all metrics for both YouTubeMatte resolutions, low-resolution VideoMatte240K, and HHM2K. On VideoMatte240K at 1080p, SnapMat achieves the lowest MSE, Grad, and dtSSD, while MatAnyone 2 yields slightly lower MAD and Conn. On VM4K, SnapMat achieves the lowest MAD, MSE, Conn, and dtSSD, while MatAnyone 2 obtains a lower Grad. Figure~\ref{fig:qualitative-comparison} further illustrates the recovery of fine structures such as hair strands and racket meshes, together with the complete foreground extent of associated objects.\par\endgroup

\subsection{Inference Efficiency and GPU Memory}

\begingroup\tolerance=1800\looseness=-1
Table~\ref{tab:efficiency} reports steady-state FPS and peak allocated GPU memory without model compilation. On the RTX 5090, SnapMat achieves near-real-time 1080p and 4K processing at 24.16 and 21.76 FPS, versus 7.86 and 2.40 FPS for MaGGIe + SAM2. On the RTX 4090, SnapMat also sustains 16.57 FPS at 4K, compared with 1.69 FPS for MaGGIe + SAM2. Memory growth is modest on both GPUs: from 1080p to 4K, SnapMat increases from 1.78 to 2.11 GB, compared with 4.65 to 14.50 GB for MaGGIe + SAM2. MatAnyone 2 reaches 17.35 GB at 1440p and runs out of memory at 4K. SnapMat thus maintains high-resolution throughput with substantially lower memory usage. Importantly, SnapMat achieves this efficiency with native-resolution refinement. Timing details are in the appendix.\par\endgroup

\subsection{Ablations}

\noindent\textbf{Module composition.} On native-resolution HHM2K, we compare backbone-only, DICE-only, SHARP-only, and full-model variants with matched initialization and update counts. DICE-only uses pointwise mappings instead of SHARP feature blocks, retaining the shared $3\times3$ sparse residual head. SHARP-only replaces DICE with feature concatenation and $1\times1$ convolution. Both single-module variants retain the same region selection, \mbox{residual composition, and supervision}.

\begingroup
\setlength{\intextsep}{1.5pt plus 0.5pt minus 0.5pt}
\begin{table}[H]
\centering\fontsize{7.8}{9.2}\selectfont
\caption{Module ablation on HHM2K at native resolution. Best values for each metric are shown in bold.}
\label{tab:modules}
\setlength{\tabcolsep}{1.8pt}\renewcommand{\arraystretch}{0.98}
\setlength{\aboverulesep}{0.5pt}
\setlength{\belowrulesep}{0.5pt}
\begin{tabular*}{\linewidth}{@{\extracolsep{\fill}}cccccc@{}}
\toprule
\textbf{DICE} & \textbf{SHARP} & \NBmetric{MAD} & \NBmetric{MSE} & \NBmetric{Grad} & \NBmetric{Conn} \\
\midrule
 & & \NBplain{3.42} & \NBplain{0.88} & \NBplain{29.35} & \NBplain{42.08} \\
\NBenabled & & \NBplain{3.35} & \NBplain{0.85} & \NBplain{28.50} & \NBplain{40.46} \\
 & \NBenabled & \NBplain{2.95} & \NBplain{0.74} & \NBplain{17.96} & \NBplain{33.56} \\
\NBenabled & \NBenabled & \NBbold{2.89} & \NBbold{0.71} & \NBbold{17.41} & \NBbold{32.64} \\
\bottomrule
\end{tabular*}
\end{table}

\endgroup

As shown in Table~\ref{tab:modules}, the full model achieves the best results on all four metrics, reducing MAD and Grad by 15.57\% and 40.70\%, respectively, relative to the backbone-only model. SHARP provides the main gains, and DICE further reduces errors, indicating that sparse correction and context fusion are complementary.

\noindent\textbf{Input resolution.} We fix the full-model checkpoint and compare refinement disabled, refinement with the input long side capped at 1920 pixels, and native-resolution refinement. All predictions are resized to the annotation dimensions for scoring. The coarse prediction in Table~\ref{tab:resolution} comes from the full-model checkpoint, not the independently trained backbone-only configuration in Table~\ref{tab:modules}.

\begingroup
\setlength{\intextsep}{1.5pt plus 0.5pt minus 0.5pt}
\begin{table}[H]
\centering\fontsize{7.8}{9.2}\selectfont
\captionsetup{justification=raggedright,singlelinecheck=false}
\caption{Coarse and refined outputs on HHM2K and VM240K.}
\label{tab:resolution}
\setlength{\tabcolsep}{1.5pt}\renewcommand{\arraystretch}{1.12}
\setlength{\aboverulesep}{0.5pt}
\setlength{\belowrulesep}{0.5pt}
\begin{tabular*}{\linewidth}{@{\extracolsep{\fill}}llccccc@{}}
\toprule
\textbf{Set} & \textbf{Output} & \NBmetric{MAD} & \NBmetric{MSE} & \NBmetric{Grad} & \NBmetric{Conn} & \NBmetric{dtSSD} \\
\midrule
HHM2K & Coarse & \NBplain{3.47} & \NBplain{0.88} & \NBplain{28.15} & \NBplain{42.09} & \NBmissing \\
 & Refined (1920) & \NBplain{3.19} & \NBplain{0.79} & \NBplain{22.44} & \NBplain{39.25} & \NBmissing \\
 & Refined (native) & \NBbold{2.89} & \NBbold{0.71} & \NBbold{17.41} & \NBbold{32.64} & \NBmissing \\
\midrule
VM240K & Coarse & \NBplain{4.18} & \NBplain{0.21} & \NBplain{3.07} & \NBplain{1.63} & \NBplain{1.11} \\
 & Refined & \NBbold{4.08} & \NBbold{0.21} & \NBbold{2.88} & \NBbold{1.52} & \NBbold{1.11} \\
\bottomrule
\end{tabular*}
\end{table}
\endgroup

Table~\ref{tab:resolution} shows that both refinement settings reduce all four errors on HHM2K. Native resolution further reduces MAD by 9.32\% and Grad by 22.41\% relative to the 1920-pixel limit, showing that preserving native image details improves refinement. The main HHM2K comparison retains the common 1920-pixel input limit. On VM240K videos, refinement also reduces MAD, Grad, and Conn, showing that the branch improves alpha prediction in video frames.

\subsection{Data Effectiveness}

\begingroup\tolerance=3000\looseness=-1
\noindent\textbf{Gains for the full model.} With the SnapMat architecture and number of training updates fixed, we vary \mbox{data composition} and sampling proportions to compare existing data, the addition of REALM images, and the further addition of REALM videos (Table~\ref{tab:data-ablation}).\par\endgroup

\begingroup
\setlength{\intextsep}{1.5pt plus 0.5pt minus 0.5pt}
\begin{table}[H]
\centering
\caption{Data composition ablation for SnapMat with a fixed update budget. I and V denote REALM-Image10K and REALM-Video1K.}
\label{tab:data-ablation}
\fontsize{8}{9.6}\selectfont
\setlength{\tabcolsep}{2.1pt}
\renewcommand{\arraystretch}{1.08}
\setlength{\aboverulesep}{0.5pt}
\setlength{\belowrulesep}{0.5pt}
\begin{tabular*}{\linewidth}{@{\extracolsep{\fill}}lccccc@{}}
\toprule
\textbf{Training data} & \textbf{MAD}$\downarrow$ & \textbf{MSE}$\downarrow$ & \textbf{Grad}$\downarrow$ & \textbf{Conn}$\downarrow$ & \textbf{dtSSD}$\downarrow$ \\
\midrule
\multicolumn{6}{@{}l@{}}{\textit{YouTubeMatte ($1920\times1080$)}} \\
\midrule[0.3pt]
Existing & 1.47 & 0.47 & 3.76 & 2.64 & 1.23 \\
Existing + I & 1.39 & 0.39 & 3.80 & 2.49 & 1.24 \\
Existing + I + V & \textbf{1.27} & \textbf{0.28} & \textbf{3.58} & \textbf{2.22} & \textbf{1.22} \\
\midrule
\multicolumn{6}{@{}l@{}}{\textit{HHM2K (native resolution)}} \\
\midrule[0.3pt]
Existing & 3.08 & 0.79 & 19.33 & 35.35 & -- \\
Existing + I & 2.90 & 0.72 & 17.74 & 32.72 & -- \\
Existing + I + V & \textbf{2.89} & \textbf{0.71} & \textbf{17.41} & \textbf{32.64} & -- \\
\bottomrule
\end{tabular*}
\end{table}

\endgroup

\begingroup\tolerance=3000\looseness=-1

On YouTubeMatte, adding only images reduces MAD, MSE, and Conn but slightly increases Grad and dtSSD. Further adding videos improves all five metrics over both preceding configurations. Relative to existing data alone, the full configuration reduces MAD, MSE, and dtSSD by 13.50\%, 39.55\%, and 1.24\%, respectively. On HHM2K, generated images provide most of the gains, while videos offer smaller additional improvements, indicating different roles for the two types of supervision across tasks.\par\endgroup

\begingroup\tolerance=1800\looseness=-1
\noindent\textbf{Four-frame video construction.} To control source-frame counts, we compare 40 VM240K videos with 100 frames each against 1,000 four-frame REALM clips, giving 4,000 source frames in each group. Both groups start from the same SAM2.1 Hiera-L pretrained weights and train for 18,000 updates under matched settings using only the backbone. VM240K videos are divided into non-overlapping four-frame clips. Both groups sample two frames from each clip and provide a target mask only for the \mbox{first sampled frame}.\par\endgroup

\begingroup
\setlength{\intextsep}{2.5pt plus 1pt minus 0.5pt}
\begin{table}[H]
\centering
\caption{Video-data comparison on YouTubeMatte using the SAM2 backbone, with 4,000 source frames per group.}
\label{tab:video-data}
\fontsize{8}{9.6}\selectfont
\setlength{\tabcolsep}{1.8pt}
\renewcommand{\arraystretch}{1.20}
\setlength{\aboverulesep}{1.0pt}
\setlength{\belowrulesep}{1.0pt}
\begin{tabular*}{\linewidth}{@{\extracolsep{\fill}}lccccc@{}}
\toprule
\textbf{Training data} & \textbf{MAD}$\downarrow$ & \textbf{MSE}$\downarrow$ & \textbf{Grad}$\downarrow$ & \textbf{Conn}$\downarrow$ & \textbf{dtSSD}$\downarrow$ \\
\midrule
\multicolumn{6}{@{}l@{}}{\textit{YouTubeMatte ($512\times288$)}} \\
\midrule[0.3pt]
VM240K & 5.41 & 0.61 & 0.57 & 0.34 & 1.25 \\
REALM-Video1K & \textbf{2.39} & \textbf{0.41} & \textbf{0.39} & \textbf{0.26} & \textbf{1.19} \\
\midrule
\multicolumn{6}{@{}l@{}}{\textit{YouTubeMatte ($1920\times1080$)}} \\
\midrule[0.3pt]
VM240K & 4.82 & 0.51 & 5.85 & 3.77 & 1.41 \\
REALM-Video1K & \textbf{1.70} & \textbf{0.32} & \textbf{4.44} & \textbf{2.79} & \textbf{1.33} \\
\bottomrule
\end{tabular*}
\end{table}

\endgroup

As shown in Table~\ref{tab:video-data}, with matched source-frame counts and update budgets, REALM-Video1K outperforms the VM240K control on all metrics at both YouTubeMatte resolutions, reducing dtSSD by 4.77\% and 5.75\%, respectively. This supports the effectiveness of generated four-frame clips under the current training protocol.

  \endgroup
  \section{Conclusion}\label{sec:conclusion}

\looseness=-1
This work demonstrates the effectiveness of separating temporal propagation from native-resolution detail recovery. SnapMat retains target and temporal context on the backbone grid and uses two-dimensional sparse refinement to correct alpha in selected regions, balancing matting accuracy, inference speed, and GPU memory usage across multiple evaluation settings. Controlled experiments on REALM further show that generated images and four-frame clips constructed through our pipeline provide effective supplementary supervision. These samples cover fine structures, semitransparent materials, and scenes in which people and designated associated objects jointly form the foreground. Sparse refinement enables efficient use of native image details, while generated images and short clips provide supplementary supervision for learning these details. Together, they support high-resolution video matting.

\fi
\par\penalty-10000
\clearpage
\setlength{\smHalfWidth}{\columnwidth}
\smRestoreLayout
\graphicspath{{figures/}}
\captionsetup{font=small,skip=10pt}
\captionsetup[figure]{skip=10pt}
\captionsetup[table]{position=top,font={small,rm},skip=3pt,justification=justified,singlelinecheck=false}
\appendix
\counterwithin{algorithm}{section}
\counterwithin{figure}{section}
\counterwithin{table}{section}
\renewcommand{\thealgorithm}{\thesection\arabic{algorithm}}
\renewcommand{\thefigure}{\thesection\arabic{figure}}
\renewcommand{\thetable}{\thesection\arabic{table}}

\renewcommand{\theHfigure}{supp.\thesection.\arabic{figure}}
\renewcommand{\theHtable}{supp.\thesection.\arabic{table}}
\setcounter{equation}{0}
\renewcommand{\theHequation}{supp.\arabic{equation}}

\pdfbookmark[0]{Appendix}{appendix-start}%
{\large\bfseries\noindent\makebox[\linewidth][l]{Appendix}\par}\nobreak\vspace{12pt}
This appendix describes the implementation and training of SnapMat
(Sec.~\ref{sec:impl}), evaluation protocols and qualitative results
(Sec.~\ref{sec:eval}), REALM data construction and review
(Sec.~\ref{sec:realm}), and inference efficiency (Sec.~\ref{sec:eff}).
\let\smOriginalSection\cvprsection
\patchcmd{\cvprsection}{-10pt plus -2pt minus -2pt}{-9pt}{}{}
\section{Implementation details}\label{sec:impl}
\let\cvprsection\smOriginalSection
\subsection{Network architecture}
DICE encodes context across the frame, while SHARP reads it at selected sites (Table~\ref{tab:proof-A1}). The gate combines pixels with intermediate coarse alpha values and the contour of the thresholded foreground. This includes sharp boundaries even where the coarse matte changes directly from foreground to background. SHARP also uses features from a narrow margin around the gate, but only predictions inside the gate modify the matte.
Algorithm~\ref{alg:infer} gives the inference steps for each frame.

\par\vspace{6pt plus 5pt}\noindent\begin{minipage}{\columnwidth}
\captionsetup{type=table}\centering\begingroup\smTableNumbers
\caption[Refinement architecture.]{Refinement architecture. Only DICE uses a capped reference grid.
It uses the padded input dimensions unless the long side exceeds 1920 pixels
or the area exceeds $1920\times1088$ pixels (1080p padded to a multiple of 16).
For a $3840\times2160$ input, DICE uses a $1920\times1088$ grid and outputs
a $480\times272$ context feature map at $1/4$ scale. SHARP retains the full
$3840\times2160$ refinement grid and operates only on $\Omega$.
$F_4$, $F_8$, and $F_M$ have 48, 96, and 256 channels.
``cond.'' denotes conditioning on coarse alpha and the gate.
Feature blocks use GroupNorm and GELU; the head is zero-initialized.
See Algorithm~\ref{alg:infer}.}\label{tab:proof-A1}
\smTableBody
\setlength{\tabcolsep}{1pt}
\begin{tabular*}{\linewidth}{@{\extracolsep{\fill}}lccc@{}}
\toprule
\textbf{Layer} & \textbf{Scale} & \textbf{Operation} & \textbf{Channels} \\
\midrule
\multicolumn{4}{@{}l@{}}{\textit{DICE: dense context encoding}} \\
\midrule[0.3pt]
Encoder 1 & $1/4$ & RGB + coarse $\alpha$ & $4{\to}64$ \\
Encoder 2 & $1/8$ & Fuse $F_4$; downsample & $112{\to}192$ \\
Bottleneck & $1/16$ & Fuse $F_8$; downsample; fuse $F_M$ & $288{\to}384$; $640{\to}384$ \\
Decoder 1 & $1/8$ & Up + skip + cond. & $384{\to}192$ \\
Decoder 2 & $1/4$ & Up + skip + cond. & $192{\to}96$ \\
\midrule
\multicolumn{4}{@{}l@{}}{\textit{SHARP: sparse residual prediction}} \\
\midrule[0.3pt]
Context & $1$ & Sample + project & $96{\to}32$ \\
Local & $1$ & Project RGB, $\alpha$, $G$ & $5{\to}32$ \\
Fusion & $1$ & Concatenate & $32+32{\to}64$ \\
Block & $1$ & Two $3\times3$ SubMConv & $64{\to}64$ \\
Head & $1$ & $3\times3$ SubMConv & $64{\to}1$ \\
\bottomrule
\end{tabular*}
\endgroup

\end{minipage}\par

\begin{algorithm}[!tbp]
\begingroup
\setlength{\parindent}{0pt}
\captionsetup{position=top,font={small,rm},skip=3pt,justification=justified,singlelinecheck=false}
\algrenewcommand\algorithmicrequire{\textbf{Input:}}
\algrenewcommand\algorithmicensure{\textbf{Output:}}
\algrenewcommand\algorithmicindent{0.9em}
\algrenewcommand\alglinenumber[1]{\fontsize{7}{8}\selectfont #1:}
\newcommand{\SMfn}[1]{\operatorname{#1}}
\caption[SnapMat video inference.]{SnapMat video inference.
$\alpha^b$: coarse output; $F$: image/memory features;
$\mathcal{D}_r$: dilation; $\partial$: contour. \textit{Prepare} aligns RGB and
coarse alpha on a canvas padded to multiples of 16, the reference grid for DICE.
At active sites, \textit{Sample} reads the corresponding DICE features; SHARP
combines them with native RGB to predict residuals. \textit{Scatter} fills
other sites with zeros; \textit{Resize} restores the output size. \textit{SAM2Step} uses
$1024\times1024$ inputs. \textbf{Only the coarse path updates memory.}}\label{alg:infer}
\smTableBody
\hrule height \heavyrulewidth
\vspace{2pt}
\begin{algorithmic}[1]
\Require Frames $I_{1:T}$, mask $P$, refinement flag $e$
\Ensure Alpha mattes $A_{1:T}$
\State $M\gets\varnothing$
\For{$t=1,\ldots,T$}
  \State $P_t\gets P$ if $t=1$, else $\varnothing$
  \State $(Z,F,\alpha^b,M)\gets\SMfn{SAM2Step}(I_t,P_t,M)$
  \If{$e=\mathrm{false}$}
    \State $A_t\gets\alpha^b$; \textbf{continue}
  \EndIf
  \State $(X,\alpha^c)\gets\SMfn{Prepare}(I_t,\sigma(Z))$
  \State $U\gets\mathbf{1}[\,0.05<\alpha^c<0.95\,]$
  \State $G\gets\mathcal{D}_4\!\left(U\cup\partial\mathbf{1}[\,\alpha^c\geq0.5\,]\right)$
  \State $\Omega\gets\SMfn{supp}(\mathcal{D}_2(G))$
  \State $D\gets\SMfn{DICE}(X,\alpha^c,F,G)$
  \State $C\gets\SMfn{Sample}(D,\Omega)$
  \State $L\gets\SMfn{Gather}([X,\alpha^c,G],\Omega)$
  \State $\delta_\Omega\gets\SMfn{SHARP}(C,L)$
  \State $\delta\gets\SMfn{Scatter}(\delta_\Omega,\Omega)$
  \State $\alpha^r\gets\SMfn{clip}(\alpha^c+G\odot\delta,0,1)$
  \State $d\gets\SMfn{Unpad}(\alpha^r-\alpha^c)$
  \State $A_t\gets\SMfn{clip}(\alpha^b+\SMfn{Resize}(d),0,1)$
\EndFor
\State \Return $A_{1:T}$
\end{algorithmic}
\vspace{1pt}
\hrule height \heavyrulewidth
\endgroup

\end{algorithm}

\subsection{Training configuration}
\noindent\textbf{Training stages.}
We initialize SAM2.1 Hiera-L from public weights~\cite{ref11} and follow the three stages in Sec.~3.4 of the main paper. We stop gradients at the memory features. All stages use AdamW on eight 24\,GB RTX 4090 GPUs with a global batch of eight frame pairs.
Table~\ref{tab:proof-A2} lists the hyperparameters for each stage.
\par\smallskip\noindent\begin{minipage}{\columnwidth}
\captionsetup{type=table}\centering\begingroup\smTableNumbers
\caption[Training hyperparameters.]{Training hyperparameters. LR: initial/peak value; --: not optimized. In joint training,
SAM2~rates apply to the trunk and neck/mask decoder.}\label{tab:proof-A2}
\smTableBody

\begin{tabularx}{\linewidth}{@{}l *{3}{J}@{}}
\toprule
\textbf{Setting} & \textbf{Adapt.} & \textbf{Refine} & \textbf{Joint} \\
\midrule
Updates & 32,400 & 30,000 & 10,000 \\
LR: encoder & $3\times10^{-5}$ & -- & $2\times10^{-7}$ \\
LR: other SAM2 & $5\times10^{-5}$ & -- & $10^{-6}$ \\
LR: refiner & -- & $5\times10^{-4}$ & $2\times10^{-5}$ \\
Weight decay & $0.1$ & $10^{-4}$ & $10^{-4}$ \\
Grad. clip: SAM2 & $0.1$ & -- & $0.1$ \\
Grad. clip: refiner & -- & $4.5$ & $4.5$ \\
Warm-up steps & $0$ & $500$ & $300$ \\
\bottomrule
\end{tabularx}
\endgroup

\end{minipage}\par\nobreak\vspace{12pt plus 6pt}

\noindent\textbf{Training data.}
For backbone adaptation, the video sources are VideoMatte240K~\cite{ref04}, CRGNN~\cite{ref23}, and
VM108~\cite{ref43}. The image sources are RefMatte~\cite{refMatte}, AM-2K~\cite{ref29},
AIM-500~\cite{aim500}, Adobe Image Matting~\cite{ref01}, PhotoMatte85~\cite{ref04},
Transparent-460~\cite{ref42}, RealWorldPortrait-636~\cite{ref03}, PPM-100~\cite{ref02},
Human2K~\cite{ref41}, P3M-10K~\cite{ref28}, and the portrait subset of Distinctions-646~\cite{ref27}.
For refinement training and joint optimization, we use images from
RealWorldPortrait-636~\cite{ref03}, Human2K~\cite{ref41}, P3M-10K~\cite{ref28},
the portrait subset of Distinctions-646~\cite{ref27}, AM-2K~\cite{ref29}, and
our REALM-Image10K (Sec.~\ref{sec:realm}), and videos from VideoMatte240K~\cite{ref04},
CRGNN~\cite{ref23}, and our REALM-Video1K (Sec.~\ref{sec:realm}).

\noindent\textbf{Sampling and training inputs.}
Each training pair consists of two frames from one video or two copies of a still image and its alpha matte. We sample 14,400 pairs per epoch for backbone adaptation and 16,000 in each subsequent stage. For online compositing, we use BG20K~\cite{ref29} backgrounds except for VM108~\cite{ref43}, which supplies its own. REALM images retain their stored composites.

Both frames undergo the same flip, scaling, crop, and color perturbation. Backbone adaptation also uses affine and grayscale transforms. For refinement and joint training, we crop to $1024\times1024$, $864\times1152$, $832\times1248$, or $720\times1280$ and also use the three rectangular crops with width and height interchanged. Training uses BF16 except for sparse convolutions and refinement losses, which use FP32.

\noindent\textbf{Losses and supervision regions.}
In Eqs.~(5) and~(6) of the main paper, the ground-truth matte defines the transition region for backbone adaptation ($0.05<\alpha^*<0.95$). If the region or its complement is empty, its contribution is zero. In the two later stages, we compute structural errors on valid image content and average them over the supervision region, or over the entire valid image if that region is empty. Padding is excluded throughout.

The Laplacian term sums absolute errors at five levels with weights of 1, 2,
4, 8, and 16, from fine to coarse. Backbone adaptation uses a $5\times5$
Gaussian smoothing kernel ($\sigma=1$). The two later stages use the outer
product of $[1,4,6,4,1]/16$ with itself. The gradient term sums the horizontal and vertical
mean absolute differences in unnormalized $3\times3$ Sobel responses.

\noindent\textbf{Training cost.}
Backbone adaptation takes about 6\,h, refinement 3\,h\,4\,min, and joint optimization 1\,h\,59\,min: 11\,h\,3\,min in total (88.4 GPU-hours).

\section{Evaluation protocols and results}\label{sec:eval}
\subsection{Evaluation data and scoring protocol}
We prompt each video only on the first frame and each still image independently. For YouTubeMatte~\cite{ref08} and VideoMatte240K~\cite{ref04}, we use the supplied RGB videos, alpha mattes, and official first-frame masks. At both $512\times288$ and $1920\times1080$, we evaluate the same 64 YouTubeMatte sequences and 50 VideoMatte240K sequences, taking 100 consecutive frames from each.

\noindent\textbf{VM4K.}
We composite each of five VideoMatte240K~\cite{ref04} test foregrounds with five fixed
BG20K~\cite{ref29} backgrounds to obtain 25 videos of 100 frames each. Each background
is resized with its aspect ratio preserved, then center-cropped to match
the foreground. We threshold the first ground-truth matte at $\alpha>0.5$
and apply morphological closing with a $15\times15$ elliptical kernel to
obtain the target mask. MatAnyone~2~\cite{ref09} runs at $2560\times1440$;
we restore its prediction to 4K for scoring.

\noindent\textbf{HHM2K.}
We evaluate all 2,000 test images~\cite{ref16} independently. We threshold the ground-truth matte at $\alpha>0$, reduce each mask dimension to one quarter of its original size, and restore the mask with nearest-neighbor interpolation. A morphological closing with a $15\times15$ elliptical kernel gives the target mask.
Table~\ref{tab:proof-B1} lists the input settings for each method.
\begin{table}[!htbp]\centering\begingroup\smTableNumbers
\caption[Inputs for quantitative evaluation.]{Inputs for quantitative evaluation. At native resolution, the largest HHM2K images exceed GPU memory for MatAnyone~2 and MaGGIe + SAM2. We therefore cap the long side at 1920 pixels for all variable-size inputs on HHM2K and score all methods at the original annotation size. SnapMat uses its coarse output at $512\times288$. MatAnyone~2 runs 10 extra iterations on the first frame. MaGGIe uses SAM2-propagated masks in three-frame windows.}\label{tab:proof-B1}
\smTableBody

\begin{tabularx}{\linewidth}{@{}l *{2}{Z}@{}}
\toprule
\textbf{Method} & \textbf{Video input} & \textbf{HHM2K input} \\
\midrule
SnapMat & Native RGB & Long side $\leq1920$ \\
SAM2Matting~\cite{ref15} & Official protocol & 1008\textsuperscript{2} \\
MatAnyone 2 & Official protocol & Long side $\leq1920$ \\
MaGGIe~\cite{ref17} + SAM2 & Official protocol & Long side $\leq1920$ \\
\bottomrule
\end{tabularx}
\endgroup
\end{table}

\begin{figure}[!b]\input{plate_b1}\end{figure}

\begin{figure*}[!tp]
\centering
\begin{minipage}{\linewidth}
\centering\normalsize
\setlength{\parskip}{0pt}
\newcommand{\panelname}[1]{\makebox[.19231\linewidth][c]{\smSameLabel{\bfseries #1}{#1}}}
\noindent\panelname{Input RGB}\hfill\panelname{Coarse alpha}\hfill
\panelname{Gate overlay}\hfill\panelname{Signed correction}\hfill
\panelname{Refined alpha}\par\vspace{3pt}
\includegraphics[width=\linewidth]{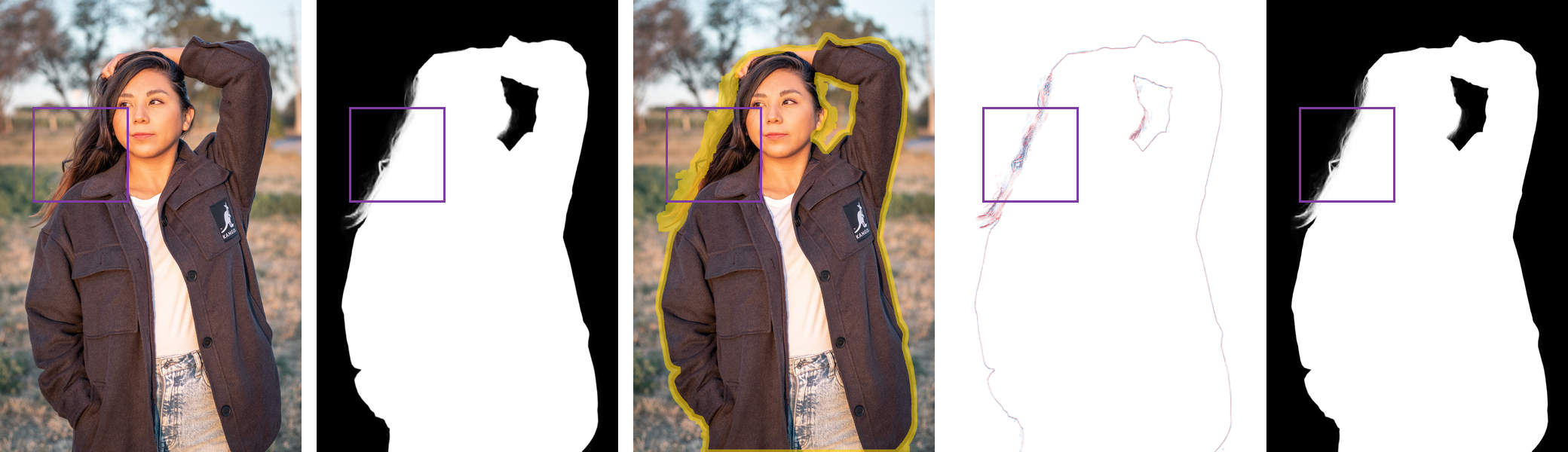}\par
\vspace{3pt}\smSameLabel{Matched detail crops ($3.2\times$)}{Matched detail crops (3.2\char128)}\par\vspace{3pt}
\includegraphics[width=\linewidth]{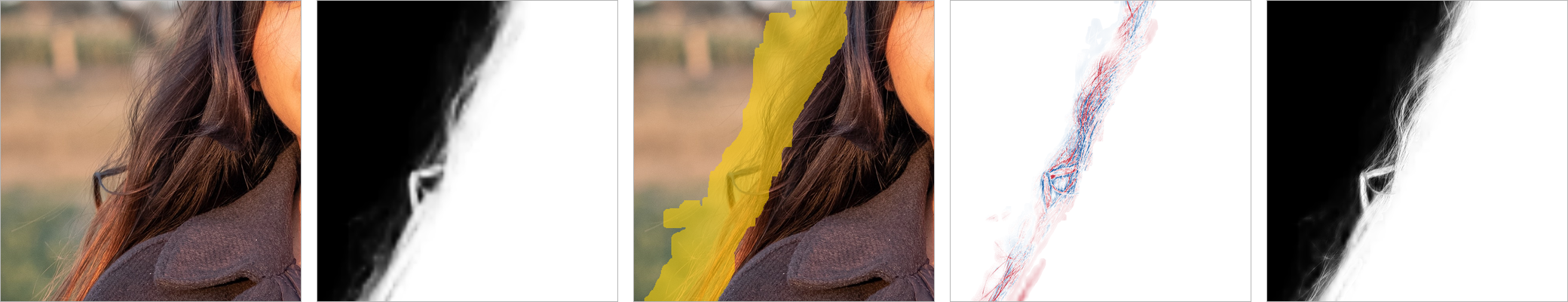}\par\vspace{4pt}
\makebox[.27\linewidth][r]{\textcolor[RGB]{38,103,171}{\smLiteral{Alpha decreases}}}\hfill
\includegraphics[width=.32\linewidth,height=6pt]{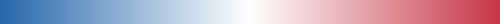}\hfill
\makebox[.27\linewidth][l]{\textcolor[RGB]{183,43,60}{\smLiteral{Alpha increases}}}\par
\makebox[.32\linewidth]{\makebox[0pt]{\smSameLabel{$\leq-0.25$}{\char129\kern.15em\char131 0.25}}\hfill\smSameLabel{$0$}{0}\hfill\makebox[0pt]{\smSameLabel{$\geq+0.25$}{\char130\kern.15em+0.25}}}
\end{minipage}
\caption{From coarse prediction to sparse refinement. From left to right: input RGB,
coarse alpha, gate overlay, signed output correction, and refined alpha.
Yellow marks the gate. Red and blue indicate increases and decreases in alpha, respectively.
The lower row enlarges the same purple-boxed region by $3.2\times$ across all five panels.
The signed correction, $\Delta\alpha=\alpha_{\mathrm{refined}}-\alpha_{\mathrm{coarse}}$,
is the change in output alpha after refinement.}
\label{fig:process}
\end{figure*}

\noindent\textbf{Scoring.}
We clip alpha to $[0,1]$ and score the full annotation grid (Sec.~5.2 of the main paper; Conn~\cite{ref51}). We average within videos, then weight videos equally; HHM2K images also carry equal weight. The first frame counts for spatial metrics and contributes zero to dtSSD without reducing the frame count. Figure~\ref{fig:gate} averages the share of coarse error inside gate $G_i$ over images with nonzero error:
\[
C_i=\frac{\sum_{p\in G_i} e_{i,p}}{\sum_p e_{i,p}},
\qquad e_{i,p}=\lvert\alpha^c_{i,p}-\alpha^*_{i,p}\rvert .
\]

\subsection{Gate coverage and refinement gains}
We compare the coarse and refined mattes from the same forward pass.
After unpadding and alignment to the annotation, we compute gate area
and within-gate MAD per image, then average.

On HHM2K~\cite{ref16},
the gate covers 5.00\% of pixels but contains 91.26\% of coarse L1 error
on average. This fraction measures where the error lies, not how much refinement removes.
Figure~\ref{fig:process} shows where refinement increases or decreases alpha within one such gate.

\subsection{High-resolution detail comparisons}
\begin{figure*}[!tp]
\noindent\begin{minipage}[t]{\columnwidth}The foreground prompt distinguishes shadow from transparency: skin and opaque fabric remain opaque in shadow. Gaps between hair strands reveal the background.\end{minipage}\hfill\begin{minipage}[t]{\columnwidth}Mixed edge pixels may have fractional alpha. A clear umbrella canopy can transmit the replacement background, whereas its ribs and handle remain opaque.\end{minipage}
\par\smallskip\centering
\includegraphics[width=\linewidth]{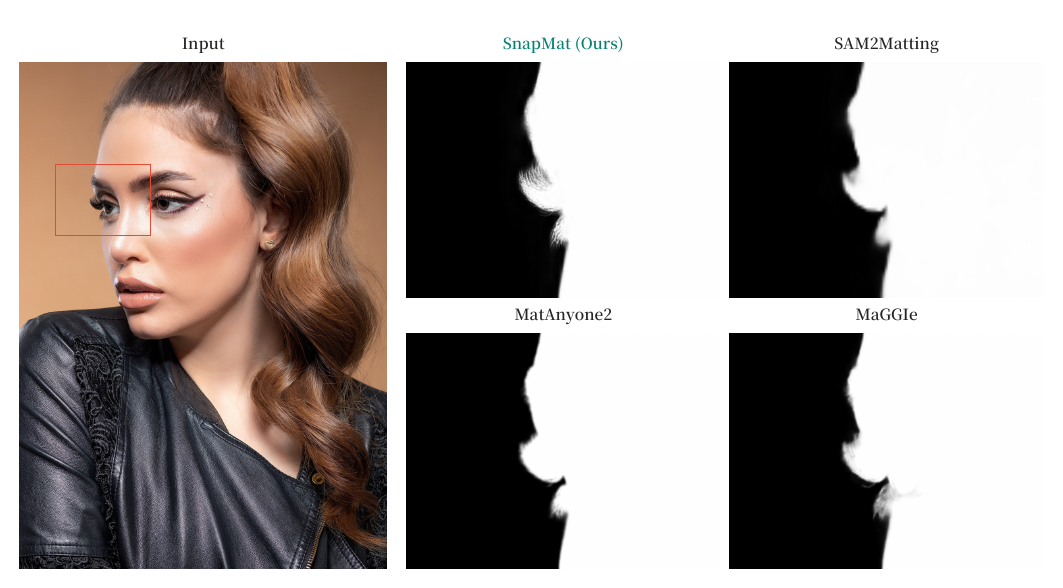}\par\smallskip
\includegraphics[width=\linewidth]{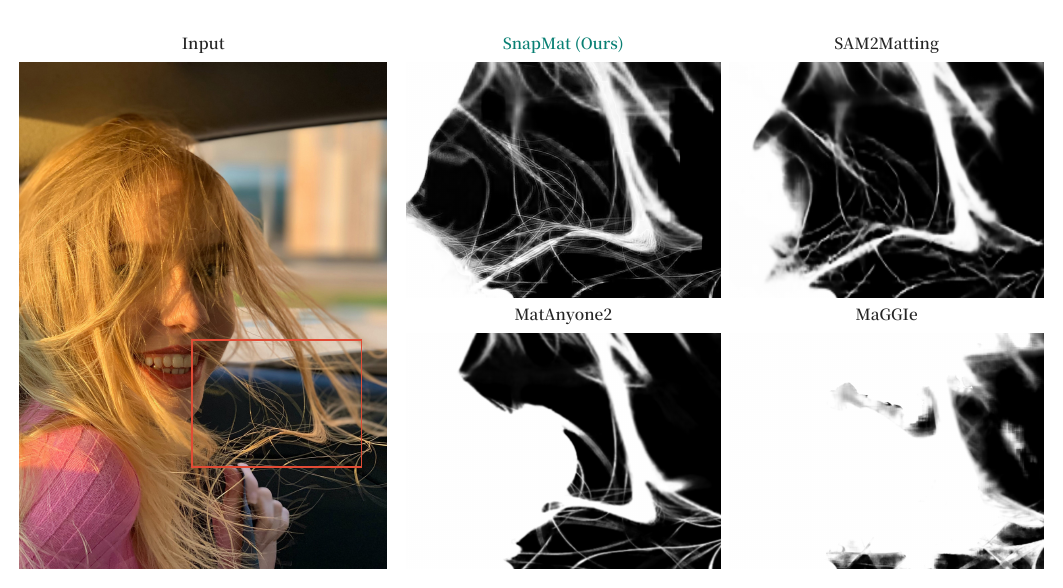}
\caption{Eyelashes (top) and flyaway hair (bottom) on HHM2K. Red boxes mark the RGB regions enlarged at right. Each grid shows alpha crops from SnapMat and SAM2Matting above those from MatAnyone~2 and MaGGIe + SAM2 (labeled MaGGIe).}
\label{fig:details1}

\end{figure*}

Figure~\ref{fig:details1} compares eyelashes and flyaway hair on two
HHM2K~\cite{ref16} images, with crop coordinates, magnification, and
grayscale range held fixed across methods.

\subsection{Input resolution and refinement details}
\begin{figure*}[p]
\centering
\resfigure{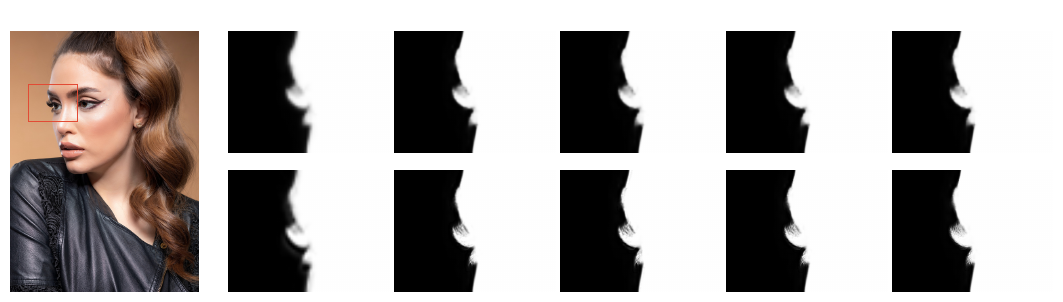}{288\times360}{1024\times1280}{1080\times1350}{1440\times1800}{2160\times2700}\par\vspace{4pt}
\resfigure{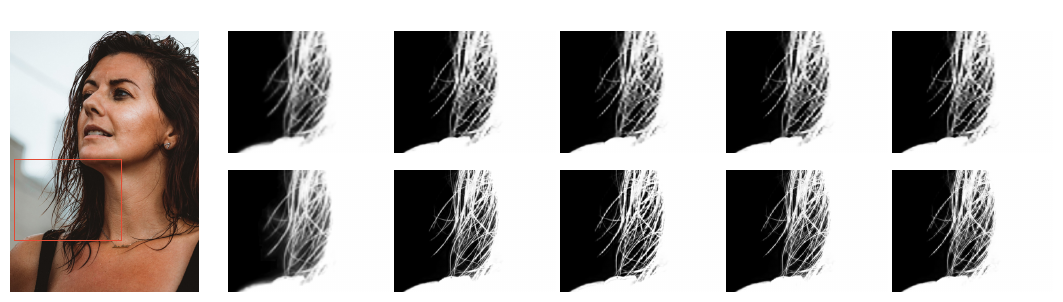}{432\times288}{1536\times1024}{1620\times1080}{2160\times1440}{3240\times2160}\par\vspace{4pt}
\resfigure{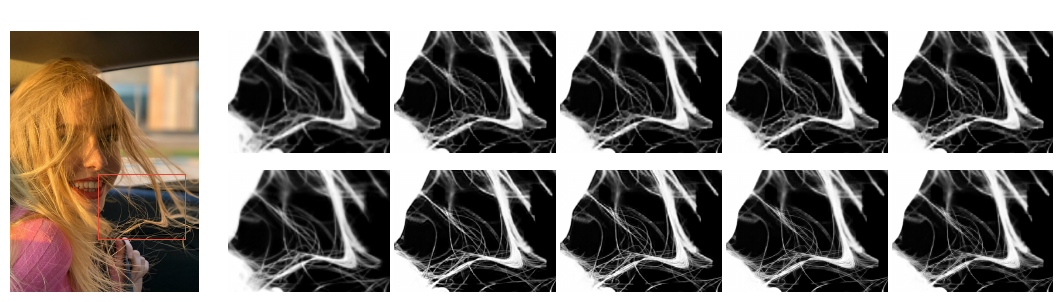}{288\times384}{1024\times1365}{1080\times1440}{1440\times1920}{2160\times2880}\par\vspace{4pt}
\resfigure{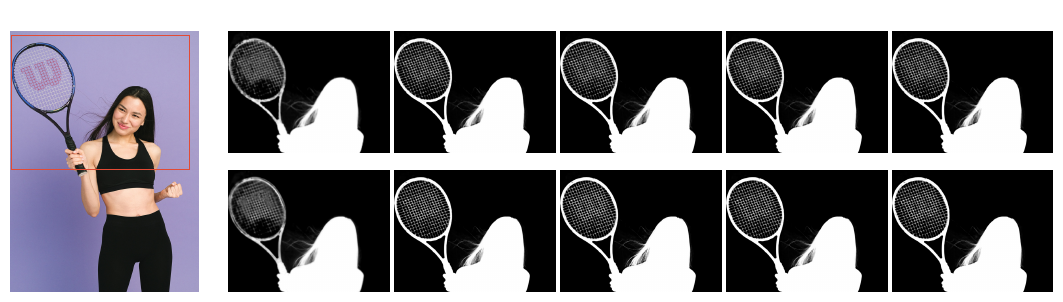}{288\times432}{1024\times1536}{1080\times1620}{1440\times2160}{2160\times3240}
\caption{Coarse and refined details at different input resolutions.
Top to bottom: eyelashes, hair tips, flyaway hair, and racket strings.
Each example has an RGB locator image on the left, with a red-boxed
region. Five columns on the right show increasing input scales,
labeled by actual width $\times$ height in pixels. Upper and lower
rows show coarse and SnapMat-refined alpha, respectively. Regions,
display sizes, and grayscale ranges are matched.}
\label{fig:multires}
\end{figure*}

Figure~\ref{fig:multires} compares coarse and refined mattes at five input
scales, mapping each crop to the same original-image region.

\subsection{Video sequence visualizations}
\begin{figure*}[p]
\centering
\begingroup
\sbox{\panelart}{\includegraphics[width=\linewidth,trim=0bp 3bp 0bp 1bp,clip]{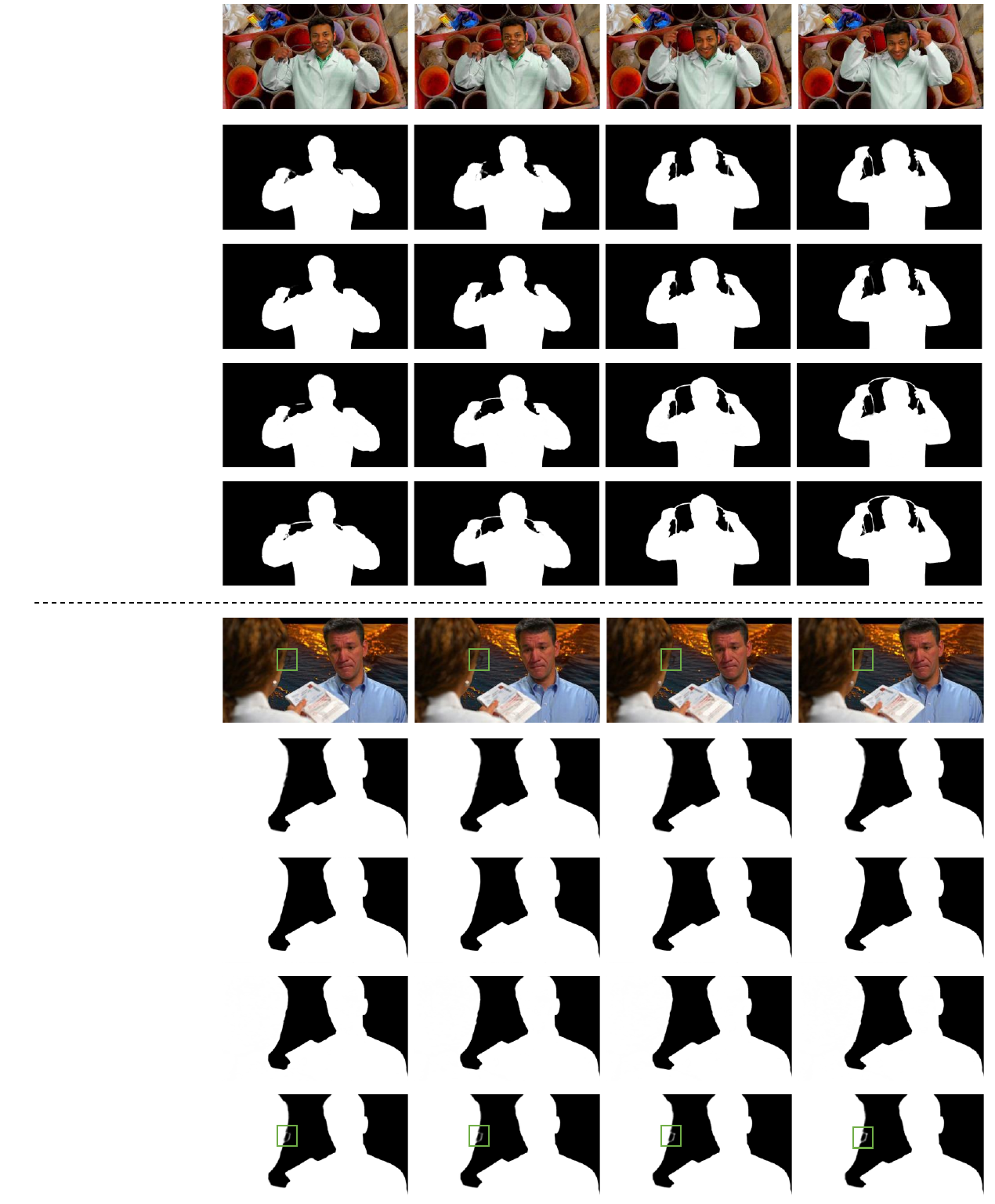}}%
\setlength{\unitlength}{\dimexpr\wd\panelart/6095\relax}%
\newcommand{\smBRow}[2]{\put(1283,#1){\makebox(0,0)[r]{#2}}}%
\newcommand{\smBPair}[2]{\shortstack[r]{{\smFigureFace #1}\\{\smFigureSubFace #2}}}%
\leavevmode\rlap{\usebox{\panelart}}%
\rlap{\smash{\begin{picture}(0,0)
\smBRow{6995}{\smBPair{Frame}{(YouTubeMatte)}}
\smBRow{6257}{{\smFigureFace MaGGIe}}
\smBRow{5524}{{\smFigureFace MatAnyone2}}
\smBRow{4799}{{\smFigureFace SAM2Matting}}
\smBRow{4074}{\smBPair{SnapMat}{(Ours)}}
\smBRow{3237}{\smBPair{Frame}{(YouTubeMatte)}}
\smBRow{2499}{{\smFigureFace MaGGIe}}
\smBRow{1766}{{\smFigureFace MatAnyone2}}
\smBRow{1041}{{\smFigureFace SAM2Matting}}
\smBRow{316}{\smBPair{SnapMat}{(Ours)}}
\end{picture}}}%
\phantom{\usebox{\panelart}}\endgroup
\caption{Sampled-frame comparisons on YouTubeMatte. The dashed line
separates two video examples; each group's four columns correspond to
four sampled time points. Rows show RGB and alpha from MaGGIe + SAM2,
MatAnyone 2, SAM2Matting, and SnapMat. The upper example shows thin structures near the head;
green boxes in the lower example mark local details near an occlusion.}
\label{fig:ytbsequence}
\end{figure*}

Figures~\ref{fig:ytbsequence} and~\ref{fig:longsequence} show sampled
predictions on YouTubeMatte~\cite{ref08} and on generated videos, respectively.

\newsavebox{\smVideoExample}
\savebox{\smVideoExample}{\begin{minipage}{\textwidth}\captionsetup{type=figure}\centering
\begingroup
  \sbox{\panelart}{\includegraphics[width=\linewidth]{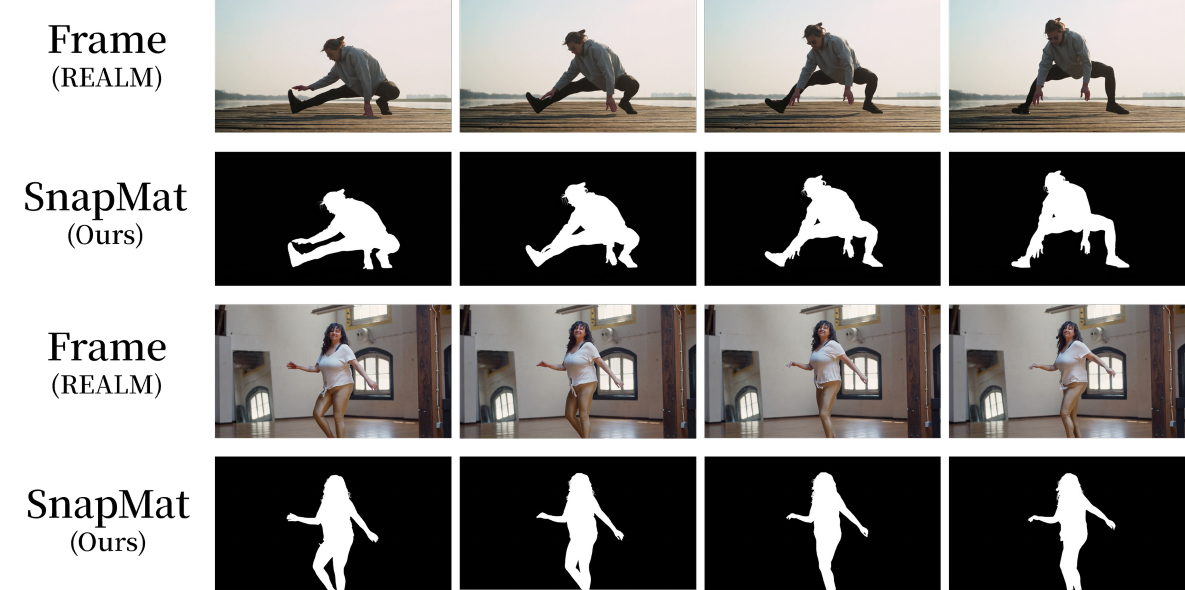}}%
  \setlength{\unitlength}{\dimexpr\wd\panelart/2605\relax}%
  \leavevmode\rlap{\usebox{\panelart}}%
  \rlap{\smash{\begin{picture}(0,0)\smMask{92}{1080}{286}{173}
\smVideoRow{235.00}{1166.50}{Frame}{(REALM)}
\smMask{40}{735}{385}{175}
\smVideoRow{235.00}{822.50}{SnapMat}{(Ours)}
\smMask{92}{406}{286}{173}
\smVideoRow{235.00}{492.50}{Frame}{(REALM)}
\smMask{46}{61}{384}{175}
\smVideoRow{235.00}{148.50}{SnapMat}{(Ours)}
\end{picture}}}%
  \phantom{\usebox{\panelart}}\endgroup
\caption{SnapMat predictions on generated 100-frame videos: outdoor
stretching (top) and indoor dancing (bottom). Each group shows four
sampled time points from left to right, with input RGB above and the
corresponding predicted alpha below. We use these videos only to illustrate inference, not for training.}
\label{fig:longsequence}
\end{minipage}}

\section{REALM data construction}\label{sec:realm}

\begin{figure*}[!tp]
\centering\usebox{\smVideoExample}\par\smallskip
\begin{minipage}[t]{\smHalfWidth}\noindent\textbf{Example prompt.}
Figure~\ref{fig:realm-prompt} shows the prompts and layers for yoga
clip \texttt{v4535143-7beedeb389-s0000}. We sample
frames at 0.08, 0.28, 0.48, and 0.68 seconds from the same shot.
The yoga mat stays in the background despite touching the subject's feet.\end{minipage}\hfill\begin{minipage}[t]{\smHalfWidth}The foreground request preserves the pose, crop, and scale in each frame;
the background request reconstructs only what the subject occludes and
preserves viewpoint and lighting. Each frame yields a $1536\times864$
transparent foreground and a matching opaque background.\end{minipage}\par\smallskip
\input{plate_c1}
\end{figure*}
\begin{figure*}[!b]\centering\input{plate_c2}\end{figure*}
\begin{figure*}[!tp]\centering\input{plate_c3}\par\medskip
\begin{minipage}{\linewidth}\captionsetup{type=figure}\centering
\includegraphics[width=\linewidth]{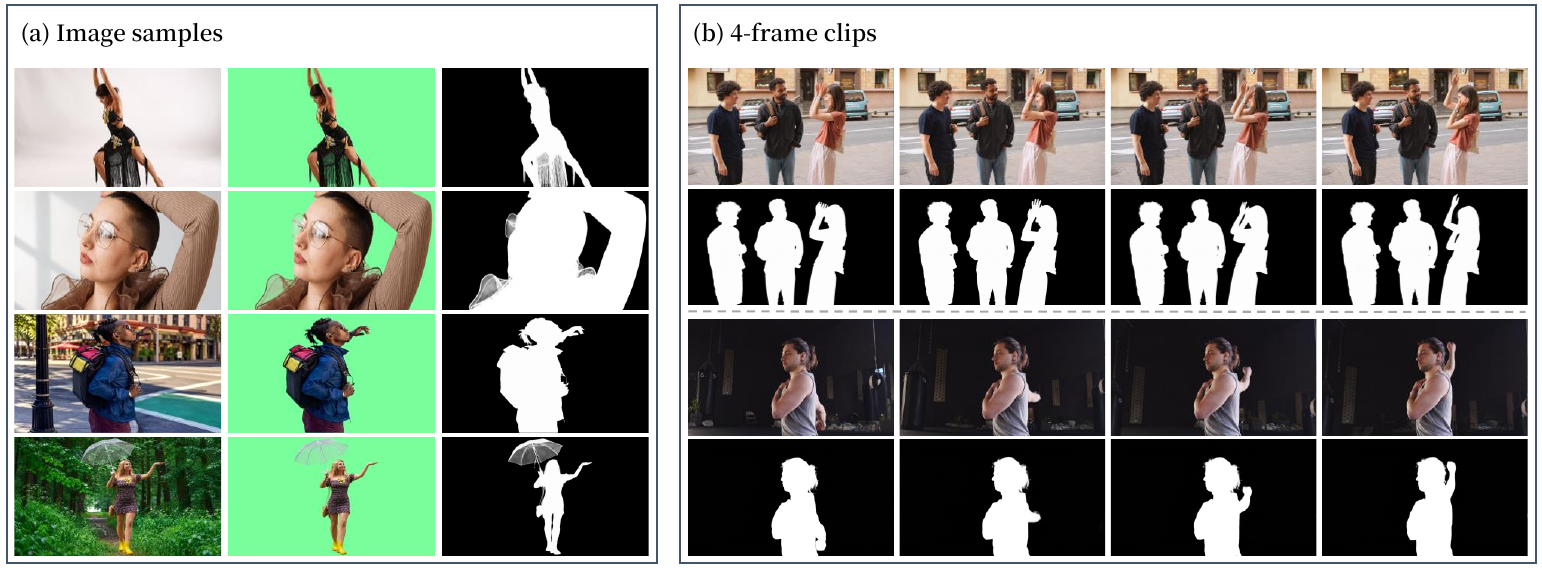}
\caption{Examples from REALM. (a) Each image is shown with its foreground composited onto green and its alpha matte. (b) Each clip shows four sampled RGB frames with their corresponding alpha mattes directly below.}
\label{fig:realm-accepted}
\end{minipage}
\end{figure*}
\subsection{Rationale and generation settings}
\noindent\begin{minipage}{\columnwidth}Table~\ref{tab:proof-C1} lists the settings for individual images and four-frame clips.\par\smallskip
\centering\captionsetup{type=table}\begingroup\smTableNumbers
\caption{Generation settings for images and videos.}\label{tab:proof-C1}
\smTableBody
\begin{tabular*}{\linewidth}{@{\extracolsep{\fill}}l@{\hspace{3pt}}c@{}c@{}}
\toprule
\textbf{Setting} & \textbf{REALM-Image10K} & \textbf{REALM-Video1K} \\
\midrule
Samples & 10,054 images & 1,007 clips (4,028 frames) \\
Generator & GPT Image 2.5 & GPT Image 2.5 or Qwen-Image-2.1 \\
Reference & One photograph & 4 same-shot frames, \({\sim}0.2\)\,s apart \\
Output size & Mostly $1536\times1024$ & 1536\,px long edge; native ratio \\
Stored layers & F,\,B,\,A, composite RGB & Same layers \\
\bottomrule
\end{tabular*}

\endgroup
\end{minipage}\par

\noindent\textbf{Layer generation.}
We generate a transparent foreground and an opaque background of the same size. For GPT Image~2.5, we request a transparent PNG and read its A channel. For Qwen-Image-2.1, \texttt{QwenImage21Pipeline} outputs RGBA through a four-channel VAE decoder (BF16, 40 steps, guidance 1.0).

We log sample sources, settings, and revisions. We use the official Pexels API for non-commercial academic research. Generation and vision-model review send material to third-party services.

\Needspace{6\baselineskip}
\noindent\textbf{Target definition.}
For each image, the target list names the person and any worn, hand-held, or attached items to retain. The ground and supporting furniture belong to the background by default. To remove exactly those targets, the background generator takes both the reference photo and the generated foreground. In video clips, the scene description and target list stay fixed across frames. We generate each frame independently: if a listed item is partly occluded, only its visible portion belongs in the foreground.

\begin{table*}[!tp]\centering\begin{minipage}{\linewidth}\captionsetup{type=table}\begingroup\smTableNumbers
\caption[Scale and organization of matting datasets.]{Scale and organization of matting datasets. HHM50K was released as a training set, but we did not use it to train SnapMat because it did not meet our data-quality requirements.
For REALM images, one reference unit is counted per sample.
$F$, $B$, $\alpha$, and $I$ denote foreground color, background, alpha, and composite RGB.}
\label{tab:realm-resources}
\smWideTableBody
\newlength{\resourcewidth}
\setlength{\resourcewidth}{\dimexpr\linewidth-10\tabcolsep\relax}
\begin{tabular}{@{}>{\raggedright\arraybackslash}m{.154\resourcewidth}
  >{\centering\arraybackslash}m{.141\resourcewidth}
  >{\centering\arraybackslash}m{.160\resourcewidth}
  >{\centering\arraybackslash}m{.194\resourcewidth}
  >{\centering\arraybackslash}m{.233\resourcewidth}
  >{\centering\arraybackslash}m{.118\resourcewidth}@{}}
\toprule
\textbf{Dataset} & \textbf{Use} & \textbf{Sources} & \textbf{Samples} &
\textbf{Construction} & \textbf{Available data}\\
\midrule
HHM50K~\cite{ref16} & Training & 50,000 images & 50,000 images &
Ultra-high-resolution portraits & RGB, $\alpha$\\
HHM2K~\cite{ref16} & Evaluation & 2,000 images & 2,000 images &
Ultra-high-resolution portraits & RGB, $\alpha$\\
VideoMatte240K~\cite{ref04} & Training, evaluation & 484 videos & 240,709 frames &
Green-screen compositing & $F$, $\alpha$\\
\midrule
\mbox{REALM-Image10K} & Training & 10,054 reference units & 10,054 images &
Photo-guided layer generation & $F$, $B$, $\alpha$, $I$\\
\mbox{REALM-Video1K} & Training & 360 videos &
1,007 clips (4,028 frames) &
Layers from same-shot frames & $F$, $B$, $\alpha$, $I$\\
\bottomrule
\end{tabular}
\endgroup
\end{minipage}\par\medskip\begin{minipage}{\linewidth}\captionsetup{type=table}\begingroup\smTableNumbers
\caption[Content coverage in REALM images and video clips.]{Content coverage in REALM images and video clips. Categories may overlap. Shares use all images or all four-frame clips as
the denominator, with each clip counted once per category. The last row
reports the recorded combined label.}
\label{tab:realm-composition}
\smWideTableBody
\begin{tabularx}{\linewidth}{@{}>{\raggedright\arraybackslash}p{.225\linewidth}
  >{\centering\arraybackslash}p{.255\linewidth} *{4}{J}@{}}
\toprule
\multirow{2}{*}[-1.2pt]{\textbf{Dimension}} & \multirow{2}{*}[-1.2pt]{\textbf{Category}} &
\multicolumn{2}{c}{\textbf{Images} ($N=10{,}054$)} &
\multicolumn{2}{c}{\textbf{Video clips} ($N=1{,}007$)}\\
\cmidrule(lr){3-4}\cmidrule(l){5-6}
& & Count & Share & Count & Share\\
\midrule
\multirow{2}{*}{Target composition} & Multiple people          & 240   & 2.39\%  & 59  & 5.86\%\\
                   & Hand-held objects        & 2,789 & 27.74\% & 422 & 41.91\%\\
\addlinespace[3pt]
\multirow{2}{*}{Interaction and occlusion} & Person--object interaction & 2,877 & 28.62\% & 450 & 44.69\%\\
                          & Occlusion by objects       & 3,084 & 30.67\% & 91  & 9.04\%\\
\addlinespace[3pt]
\multirow{2}{*}{Accessories} & Worn accessories        & 6,344 & 63.10\% & 403 & 40.02\%\\
            & Thin linear accessories & 2,750 & 27.35\% & 247 & 24.53\%\\
\addlinespace[3pt]
\multirow{3}{*}{Materials and structures} & Fine cords            & 712 & 7.08\% & 24 & 2.38\%\\
                         & Openwork fabrics      & 504 & 5.01\% & 32 & 3.18\%\\
                         & Translucent materials & 989 & 9.84\% & 29 & 2.88\%\\
\addlinespace[3pt]
Combined conditions & Interaction + accessories & 2,843 & 28.28\% & 444 & 44.09\%\\
\bottomrule
\end{tabularx}
\endgroup
\end{minipage}\end{table*}
\subsection{Multi-view and prediction-assisted review}
\noindent\textbf{Layer checks.}
Figure~\ref{fig:realmexamples} shows details that require manual inspection. We check dimensions and RGBA format, reject degenerate mattes, and verify that the A channel matches the standalone alpha and that recompositing reproduces the stored RGB. We compare the layers with the reference and target list to find missing or misplaced objects. Black and white backdrops reveal edge contamination; alpha reveals filled holes or lost translucency. The background alone exposes duplicates hidden by the foreground. We log the location, layer, severity, and proposed fix for each issue. Major defects require correction; we note but retain slight, natural-looking soft edges.

\noindent\textbf{Prediction-assisted review.}
SnapMat (trained without REALM) and MatAnyone~2 predict alpha from the same composite and mask (candidate alpha $>0.05$). At the original image size, we compare both predictions with the candidate matte. We review images in descending order of SnapMat's mean absolute difference (MAD), using the difference map to locate discrepancies and the second prediction for comparison. Reviewers inspect the layers before revising a label, focusing on translucent materials, umbrellas, people using equipment, and unresolved cases.

For video, we initialize SnapMat with the thresholded first-frame matte and predict all four frames. We rank clips by decreasing mean frame MAD, then inspect individual frames and slow playback for changes in the target, its structure, scale, or opacity.

\subsection{Correction, reinspection, and review statistics}
We correct the affected layer and match the background to any regenerated image foreground. For video, we correct only the affected frames and layers, then reinspect the entire clip.

Model predictions are candidates for manual correction, never automatic replacements. The standalone alpha must match the final RGBA foreground's A channel pixel for pixel, and the same layers must reproduce the composite as $I=\alpha F+(1-\alpha)B$. Revising alpha therefore requires updating both the A channel and the standalone matte and recompositing the image.

After initial review, we allow one regeneration per image and restrict video fixes to selected frames and layers. We discard versions that still have major defects.

We remove low-alpha remnants far from the subject, using their location
and connection to its structure to preserve faint hair, fine lines,
and translucent materials. Each edit retains its previous layer version
and review record. We recheck the modified layers, standalone alpha,
and composite (Figure~\ref{fig:quality}).

Of the 10,054 REALM images, 8,598 (85.52\%) passed initial review;
1,456 (14.48\%) have a regeneration record. All four frames of all
1,007 accepted clips were reviewed manually (100\% coverage).
For the 14,082 accepted RGB--alpha pairs, API expenditure averaged
RMB~0.099 per pair, excluding local computing and labor.

\subsection{Dataset scale and content coverage}\label{sec:realm-coverage}
Table~\ref{tab:realm-resources} distinguishes source units from released samples and lists the available layers. Table~\ref{tab:realm-composition} summarizes the content recorded in scene descriptions and supplementary visual records. Person--object interactions occur in 44.69\% of clips and 28.62\% of images. Figure~\ref{fig:realm-accepted} shows examples from both sets.

\section{Inference efficiency evaluation}\label{sec:eff}
\subsection{Single-video benchmark}\label{sec:single-video}
On both the RTX 4090 and RTX 5090, we benchmark all methods on the same 100-frame video and first-frame mask at $1024\times1024$, 1080p, 1440p, and 4K. We time inference, memory updates, and output resizing, from GPU-ready inputs to the requested alpha matte. Model loading, decoding, input preparation, and file writing are excluded.

Table~\ref{tab:proof-D1} lists the settings. SnapMat uses its coarse path at $1024\times1024$ and its full model at higher resolutions, with FP32 sparse convolutions.
\begin{table}[!htbp]\centering\begingroup\smTableNumbers
\caption{Execution settings for speed evaluation.}\label{tab:proof-D1}
\smTableBody
\begin{tabularx}{\linewidth}{@{}l Z c@{}}
\toprule
\textbf{Method} & \textbf{Processing scale} & \textbf{Precision} \\
\midrule
SnapMat & Native-resolution refinement & FP32 / BF16 \\
SAM2Matting & Fixed 1008\textsuperscript{2} & BF16 \\
MatAnyone 2 & No size limit & FP16 \\
MaGGIe + SAM2 & Native size & FP32 / BF16 \\
\bottomrule
\end{tabularx}

\endgroup
\end{table}
Each of three runs starts with a fresh video state and the first-frame mask. For every method, we measure FPS over frames 33--96 (64 frames) and report the arithmetic mean for runs two and three.

\subsection{Multi-video benchmark}
We run the full model three times on each of 50 YouTubeMatte~\cite{ref08} videos (100 frames each), upsampled from 1080p to $3840\times2160$, on the RTX 4090. Pooling the 6,400 steady-state frames from runs two and three ($50\times2\times64$) and dividing by their total processing time gives 16.73 FPS, compared with 16.57 FPS for the single video.

Table~\ref{tab:proof-D2} also reports all three runs. The last row includes all 100 frames per video: 10,000 for runs two and three and 15,000 for all three.
\begin{table}[!htbp]\centering\begingroup\smTableNumbers
\caption{SnapMat throughput and latency on 50 YouTubeMatte videos, measured on an RTX 4090. We upsample the 1080p videos to \mbox{$3840\times2160$} before inference.}\label{tab:proof-D2}
\smTableBody
\begin{tabularx}{\linewidth}{@{}l *{2}{J}@{}}
\toprule
\textbf{Metric} & \textbf{Runs 2--3} & \mbox{\textbf{All three runs}} \\
\midrule
Steady-state FPS$\uparrow$ & 16.73 & 15.39 \\
Mean latency (ms)$\downarrow$ & 59.76 & 64.97 \\
P50 latency (ms)$\downarrow$ & 59.41 & 59.80 \\
P95 latency (ms)$\downarrow$ & 61.30 & 76.64 \\
Peak allocated memory (GB)$\downarrow$ & 2.31 & 2.31 \\
Peak reserved memory (GB)$\downarrow$ & 3.28 & 3.28 \\
Pooled FPS (all 100 frames)$\uparrow$ & 16.55 & 15.28 \\
\bottomrule
\end{tabularx}

\endgroup
\end{table}

\subsection{Active-point fraction and refinement cost}\label{sec:sparse-cost}
On the RTX 4090, we profile the sparse branch and residual head without SAM2 or video memory. For each active fraction (10\% to 100\%), we sample sites at random, average latency over 15 forward passes, and record peak allocated memory in decimal GB.

\begin{figure}[!htbp]\centering
\includegraphics[width=\linewidth]{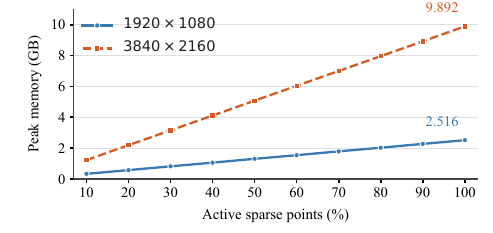}
\caption{Peak allocated memory of the sparse refinement branch as a
function of the active-point fraction at 1080p and 4K.}
\label{fig:sparsememory}
\end{figure}

\begin{figure}[!htbp]\centering
\includegraphics[width=\linewidth]{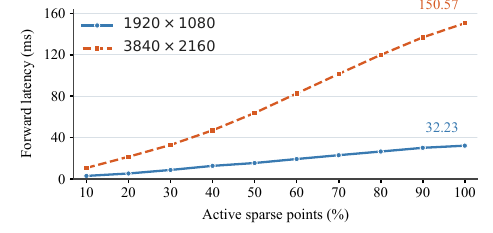}
\caption{Latency of the sparse refinement branch at 1080p and 4K. Each point is the mean of 15 forward passes at the indicated active fraction, with sites selected at random.}
\label{fig:sparselatency}
\end{figure}

From 10\% to 100\% activity, peak memory increases from 0.337 to 2.516 GB at 1080p and from 1.228 to 9.892 GB at 4K (Figure~\ref{fig:sparsememory}). The corresponding latencies rise from 3.01 to 32.23 ms and from 10.75 to 150.57 ms (Figure~\ref{fig:sparselatency}).
\Needspace{7\baselineskip}
\subsection{Module-level timing}
CUDA events on steady-state RTX 4090 frames (Table~\ref{tab:proof-D3}) show nearly constant DICE latency on its capped grid. At 4K, the image encoder takes longest; SHARP takes 5.97 ms on model-selected sites or 150.57 ms at full activity (Figure~\ref{fig:sparselatency}). With profiling overhead, 77.64 ms/frame implies 12.88 FPS versus 16.57 FPS without instrumentation.
\par\smallskip\noindent\begin{minipage}{\columnwidth}\centering\captionsetup{type=table}\begingroup\smTableNumbers
\caption[Module-level diagnostic timing of SnapMat (ms/frame).]{Module-level diagnostic timing of SnapMat (ms/frame). Intervals overlap: full refinement includes DICE and SHARP, and SHARP
includes the residual head.}\label{tab:proof-D3}
\smTableBody
\begin{tabularx}{\linewidth}{@{}l *{3}{J}@{}}
\toprule
\textbf{Module / interval} & \textbf{1080p} & \textbf{1440p} & \textbf{4K} \\
\midrule
Image encoder & 30.05 & 28.88 & 29.68 \\
Memory attention & 6.97 & 6.95 & 7.03 \\
Mask decoder & 5.98 & 6.08 & 6.20 \\
Memory encoder & 2.81 & 2.75 & 2.75 \\
Prompt encoder & 0.28 & 0.13 & 0.30 \\
\midrule
Full refinement & 21.60 & 21.36 & 24.11 \\
DICE & 12.82 & 12.13 & 12.47 \\
SHARP (incl. residual head) & 5.26 & 5.10 & 5.97 \\
Residual head & 0.61 & 0.67 & 0.61 \\
\midrule
Full-frame diagnostic interval & --- & --- & 77.64 \\
\bottomrule
\end{tabularx}
\endgroup
\end{minipage}\par

\clearpage
\pdfbookmark[0]{LLM Usage Statement}{llm-disclosure}
\section*{LLM Usage Statement}
\label{sec:llm-usage}
We used GPT Image 2.5 and Qwen Image 2.1 to generate foreground RGBA layers and backgrounds for REALM, and gpt6astra and deepseek4.1flash1017 to assist sample quality checks (Section~\ref{sec:data}). OpenAI Codex supported research discussions, code implementation and debugging, translation, language editing, proofreading, and LaTeX typesetting. Generated samples were screened and corrected using the procedures described in the paper. All reported experimental results were obtained from actual training and evaluation. The authors reviewed the AI-assisted content used in this work and take full responsibility for the scientific claims, experimental results, and references.
\par

{\flushbottom\small
\renewcommand{\bibpreamble}{\interlinepenalty=10000\relax}
\renewcommand*{\backref}[1]{\unskip\unskip\nobreakspace\mbox{#1}}
\pdfbookmark[0]{References}{all-references}
\bibliographystyle{ieeenat_fullname}
\bibliography{references}
}

\end{document}